\documentclass[review,authoryear]{elsarticle}

\usepackage{hyperref}
\usepackage{amsmath}
\allowdisplaybreaks
\usepackage{breqn}
\usepackage{calc}  
\usepackage{enumitem}  
\usepackage{amsfonts}
\usepackage{mathrsfs}
\usepackage{xcolor}
\usepackage{natbib}
\usepackage[pagewise]{lineno}
\usepackage{bigstrut}
\usepackage{booktabs}
\usepackage{rotating}
\usepackage{tabularx}
\usepackage{makecell}
\usepackage{hyperref}
\usepackage[ruled,vlined]{algorithm2e} 

\usepackage[top=3.5cm,bottom=3.5cm,left=3.5cm,right=3.5cm]{geometry}
\usepackage{caption}
\usepackage{subcaption}
\usepackage{tikz}
\modulolinenumbers[5]
\usepackage{graphicx, amsfonts,amsmath , boldline, epstopdf,amssymb }
\usepackage{tabu, multirow,multicol ,booktabs ,setspace ,algcompatible ,mathrsfs,dsfont, array, boldline, makecell, booktabs }
\usepackage{url}
\usepackage{epstopdf}
\usepackage{adjustbox, stackengine, color, colortbl, epsfig, cases, enumitem, mathtools, tikz, verbatim}
\usetikzlibrary{shapes,arrows.meta,decorations}
\definecolor{Abi}{rgb}{0.309803, 0.58039, 0.80392}
\definecolor{orange}{rgb}{1, 0.5019, 0}
\definecolor{Red}{rgb}{1, 0, 0}
\definecolor{orange}{rgb}{1, 0.50196, 0}
\definecolor{greenJ}{rgb}{0, 0.6590, 0.42}
\definecolor{Brown}{rgb}{0.588, 0.294, 0}
\definecolor{um}{rgb}{0.0824, 0.1294, 0.4196}
\definecolor{abikam}{rgb}{0.51, 0.93,  0.992}
\usepackage{amsthm}
\usepackage{lineno,hyperref}
\usepackage{amsmath}
\usepackage{breqn}
\usepackage{calc}  
\usepackage{enumitem}  
\usepackage{amsfonts}
\usepackage{mathrsfs}
\usepackage{multirow}
\usepackage{bigstrut}
\usepackage{booktabs}
\usepackage{rotating}
\usepackage{hyperref}
\usepackage{nomencl}

\makenomenclature
\renewcommand\nomgroup[1]{%
  \item[\bfseries
  \ifstrequal{#1}{I}{Sets and Indices}{%
  \ifstrequal{#1}{P}{Parameters}{%
  \ifstrequal{#1}{L}{Functions}{%
  \ifstrequal{#1}{V}{Variables}{}}}}%
]}

\usepackage{bm}
\usepackage[top=3.5cm,bottom=3.5cm,left=3.5cm,right=3.5cm]{geometry}
\usepackage{caption}
\modulolinenumbers[5]
\usepackage{mathtools}
\usepackage{stackengine}
\usepackage{lineno,hyperref}
\usepackage{amsmath}
\usepackage{breqn}
\usepackage{calc}  
\usepackage{enumitem}  
\usepackage{amsfonts}
\usepackage{mathrsfs}
\usepackage{multirow}
\usepackage{bigstrut}
\usepackage{booktabs}
\usepackage{rotating}
\usepackage{hyperref}
\usepackage[top=3.5cm,bottom=3.5cm,left=3.5cm,right=3.5cm]{geometry}

\usepackage{xcolor}
\usepackage{soul}
\usepackage{comment}

\usepackage{amsmath, amssymb}

 \usepackage{caption}
\usepackage{color}

\newcommand{\reals}{{I\kern-.35em R}}
\newcommand{\nats}{{I\kern-.35em N}}
\newcommand{\upto}{{\raise 1pt \hbox{$\scriptstyle \,\nearrow\,$}}}
\newcommand{\downto}{{\raise 1pt \hbox{$\scriptstyle \,\searrow\,$}}}
\newcommand{\eop}
	{\hfill{$\vcenter{\hrule height1pt \hbox{\vrule width1pt height5pt 
   	 \kern5pt \vrule width1pt} \hrule height1pt}$} \medskip}
\def\state #1. { \noindent{\bf#1.\enspace}}

\usepackage{lineno,hyperref}
\usepackage{amsmath}
\usepackage{breqn}
\usepackage{calc}  
\usepackage{enumitem}  
\usepackage{amsfonts}
\usepackage{mathrsfs}

\usepackage{xcolor}
\usepackage{bigstrut}
\usepackage{rotating}
\usepackage[top=3.5cm,bottom=3.5cm,left=3.5cm,right=3.5cm]{geometry}

\usepackage{subcaption}
\modulolinenumbers[5]
\usepackage{graphicx, amsfonts,amsmath , boldline, epstopdf,amssymb }
\usepackage{tabu, multirow,multicol ,booktabs ,setspace ,algcompatible ,mathrsfs,dsfont, array, boldline, makecell, booktabs }
\usepackage{url}
\usepackage{epstopdf}
\usepackage{adjustbox, stackengine, color, colortbl, epsfig, cases, enumitem, mathtools, tikz, verbatim}
\usetikzlibrary{shapes,arrows.meta,decorations}
\definecolor{um}{rgb}{0.0824, 0.1294, 0.4196}
\definecolor{Abi}{rgb}{0.059, 0.322, 0.729}
\definecolor{orange}{rgb}{1, 0.347, 0}
\definecolor{c}{rgb}{0, 1, 1}
\definecolor{m}{rgb}{1, 0, 1}

\definecolor{bb}{rgb}{0.2941, 0.5447, 0.7494}
\definecolor{greenJ}{rgb}{0, 0.6590, 0.42}
\definecolor{zereshk}{rgb}{0.588,0,0.098}
\definecolor{Maroon}{rgb}{0.502, 0, 0}
\definecolor{Brown}{rgb}{0.588, 0.294, 0}
\definecolor{Olive}{rgb}{0.502, 0.502, 0}
\definecolor{Navy}{rgb}{0, 0, 0.502}
\definecolor{Orange}{rgb}{1,0.647,0}
\definecolor{Yellow}{rgb}{0.502, 1, 0}
\definecolor{Green}{rgb}{0, 0.502,  0}
\definecolor{Blue}{rgb}{0, 0,0.761}
\definecolor{Lime}{rgb}{0.196, 0.804, 0.196}
\definecolor{Purple}{rgb}{0.502,0,0.502}
\definecolor{Violet}{rgb}{0.561,0,1}
\definecolor{Magneta}{rgb}{1,0,1}
\definecolor{Red}{rgb}{1,0,0}
\definecolor{Gray}{rgb}{0.502, 0.502, 0.502}
\definecolor{abikam}{rgb}{0.51, 0.93,  0.992}
\definecolor{abizeyad}{rgb}{0.16, 0.573,0.761}
\definecolor{rr}{rgb}{0.9047, 0.1918, 0.1988}

\definecolor{orange2}{rgb}{0.85,0.33,0.10}
\definecolor{yellow4}{rgb}{0.93,0.69,0.13}
\definecolor{purple6}{rgb}{0.49,0.18,0.56}
\definecolor{green32}{rgb}{0.47,0.67,0.19}
\definecolor{blue13}{rgb}{0.30,0.75,0.93}
\definecolor{zereshk16}{rgb}{0.64,0.08,0.18}
\definecolor{orange25}{rgb}{0.93,0.69,0.13}
\definecolor{gray27}{rgb}{0.0824, 0.1294, 0.4196}

\usetikzlibrary{tikzmark}

\tikzset{mycircled/.style={circle,draw,inner sep=0.1em,line width=0.04em}}

\tikzset{mycircled2/.style={circle,draw,inner sep=0.1em,line width=0.09em}}

 \usepackage{caption}
\journal{Transportation Research Part C: Emerging Technologies}

\begin{document}

% =========================
% Title Page
% =========================
% \begin{titlepage}
% \noindent
% \textbf{Evacuation Traffic Prediction for Dynamic Transportation Networks over Multiple Hurricanes: A Generalized Graph Learning Approach}\\

% \noindent
% Md Nafees Fuad Rafi$^{a}$, Samiul Hasan$^{a,*}$\\

% \noindent
% $^{a}$Department of Civil, Environmental and Construction Engineering, \\
% University of Central Florida, 12800 Pegasus Drive, Orlando, FL 32816, United States \\

% \noindent
% *Corresponding author: Samiul Hasan \\
% Email: \href{mailto:samiul.hasan@ucf.edu}{samiul.hasan@ucf.edu} \\

% \section*{Author Contributions Statement (CRediT)}
% \textbf{M. Rafi:} Conceptualization, Data curation, Formal analysis, Investigation, Methodology, Validation, Visualization, Writing – original draft.  
% \textbf{S. Hasan:} Conceptualization, Methodology, Funding acquisition, Resources, Supervision, Project administration, Validation, Writing – review and editing.  
% Both authors reviewed and approved the final manuscript.

% \section*{Funding}
% The study was partially supported by U.S. National Science Foundation (NSF) EAGER Grant No. 2122135. However, the authors are solely responsible for the findings presented here.

% \vfill
% \noindent

% Date: \today
% \end{titlepage}

% =========================
% Main Document Starts
% =========================

\doublespacing
\begin{frontmatter}

\title{A Dynamic Graph Learning Framework for Network-Wide Traffic Prediction during Hurricane Evacuations}


%%% Group authors per affiliation:

%\author[mythirdaddress]{Md Nafees Fuad Rafi}
%\author[mythirdaddress]{Samiul Hasan\corref{mycorrespondingauthor}}

\author{Md Nafees Fuad Rafi}
\author{Samiul Hasan\corref{mycorrespondingauthor}}

\address{Department of Civil, Environmental and Construction Engineering, University of Central Florida, 12800 Pegasus Drive, Orlando, FL 32816, United States}

\cortext[mycorrespondingauthor]{Corresponding author: samiul.hasan@ucf.edu}

%\doublespacing

\begin{abstract}
High traffic demand during hurricane evacuations causes severe congestion on major highways, leading to delays and safety risks for evacuees. Predicting evacuation traffic in real-time can significantly enhance evacuation management efforts. However, evacuation traffic is highly dynamic and uncertain due to evolving hurricane characteristics and diverse population response. Each hurricane has unique characteristics and evacuation contexts, resulting in different spatial-temporal traffic patterns. Moreover, the structure of detector-based transportation networks varies across multiple hurricanes. Hence, a model trained on the data from one hurricane may not generalize well for others. This study proposes a generalized dynamic graph learning framework using traffic detectors and evacuation data from 12 historical hurricanes. Leveraging the dynamic nature of transportation network, we develop a Graph Convolutional Long Short-Term Memory (GCN-LSTM) model that predicts evacuation traffic up to 6 hours in advance. For an unseen hurricane (Milton, 2024), the model shows strong generalizability, achieving an RMSE of 280.6 and R\textsuperscript{2} of 0.96 for 1-hour prediction, and an RMSE of 459.8 and R\textsuperscript{2} of 0.89 for overall 6-hours prediction. We also introduce a modular training approach, where we evaluate performance improvements by sequentially incorporating data from multiple hurricanes. The proposed model outperforms state-of-the-art baseline models for overall 6-hour prediction horizons. Ablation experiments show an 8.4\% improvement in RMSE for the proposed dynamic GCN-LSTM model compared to its static counterpart. Traffic management officials can use the developed model to understand future traffic conditions during hurricane evacuation in advance and take proactive measures to reduce delays during evacuation.

\end{abstract}
\begin{keyword}
Traffic Prediction \sep Generalized Graph Learning \sep Hurricane Evacuation \sep Dynamic Graph Learning \sep Hurricane Milton
\end{keyword}
\end{frontmatter}

\section{Introduction} \label{sec:intro}
% Why evacuation traffic prediction model?
Hurricanes have become more frequent and intense in the coastal regions of United States in recent times \citep{Knutson2022}, causing infrastructure damages and life-threatening risk for the residents of impacted areas. To mitigate the risks and save lives, emergency officials announce evacuation orders in different regions depending on the projected time and location of hurricane landfall. Evacuations create an abrupt surge of traffic in major highways, causing huge congestion and an increased number of crashes \citep{RAHMAN2021106260, RAHMAN2021103188}. For example, Hurricane Milton which made landfall in Florida, USA in 2024, caused an evacuation of approximately 5.5 million people from the western coast of Florida \citep{nyt2024hurricane}. It created increased congestion and crashes in the major highways of Florida, e.g. I-75, I-4, etc. \citep{ritis_incident_archive}. Traffic management officials adopt different strategies to manage high traffic congestion during evacuation such as emergency shoulder use (ESU), providing clear information, allocating contraflow and so on \citep{MURRAYTUITE201325}. However, anticipating future traffic conditions in advance can help support these evacuation traffic management strategies.

Previous studies have demonstrated the importance of advance traffic forecasting for short-term operational decision-making of agencies such as DOTs, counties, emergency management divisions. For example, \citet{HashemiAbdelghany2016} developed a real-time traffic prediction and decision-support framework and demonstrated improved network performance through various traffic management strategies in simulation experiments. Specifically for hurricane evacuations, \citet{LuEtAl2018Bottlenecks} developed a framework to identify critical bottleneck locations using evacuation demand estimates and traffic flow propagation. Furthermore, FHWA's iFlorida Model Deployment Final Evaluation Report \citep{HaasEtAl2009iFlorida} shows how predicted traffic demand during Hurricane Frances (2004) influenced the decision not to activate contraflow on I-75 in Georgia, United States. More recently, \citet{ShahEtAl2024PredictiveAnalytics} discussed how predictive analytics can support traffic management agencies in monitoring traffic conditions, identifying operational needs, and allocating resources. The report also describes applications of predictive models for positioning highway patrols and emergency response resources. Similarly, FHWA's Decision Support Framework for Dynamic Part-Time Shoulder Use \citep{FHWA2019DynamicShoulder} explains how anticipated traffic volumes can support the timely activation of shoulder use before traffic breakdown. Also, implementing emergency shoulder use and contraflow requires advance planning, field personnel, and coordination among agencies \citep{wolshon2005review1}. Reliable and more accurate advance forecasts of network-wide traffic volumes may help agencies prioritize monitoring, law enforcement, and roadside assistance resources. Therefore, a dynamic, network-wide and generalized traffic prediction model can assist transportation agencies to manage evacuation traffic more effectively.

% Challenges of evacuation traffic prediction
Evacuation traffic pattern is significantly different than traffic in regular periods due to high demand surge depending on the timings of evacuation orders and the number of people under those orders. Evacuation traffic also varies depending on hurricane trajectories and projected landfall locations, which are inherently uncertain. A sudden change in hurricane path can lead to ordering evacuations for new areas \citep{rashid2024evacuation}, consequently, changing the traffic pattern across different highways. One of the major challenges in predicting evacuation traffic is how to capture the spatiotemporal variations that occur during these critical periods. Unlike regular period traffic, where traffic generally follows predictable daily/weekly patterns, evacuation scenarios are influenced by various factors including the timing of evacuation orders \citep{RASHID2025105147}, the severity of the impending hazard, public compliance levels, and the availability of evacuation routes. These factors can lead to highly variable traffic densities, congestion hotspots, and significant fluctuations in traffic speeds, all of which should be accurately represented in the prediction model. Thus, predicting evacuation traffic is more challenging than regular traffic prediction. 

Traditionally, mathematical and simulation-based models have been used to capture evacuation traffic pattern. For instance, \cite{barrett2000dynamic} proposed a traffic modeling framework for hurricane evacuation by developing a rolling-horizon based dynamic traffic assignment (DTA) model. \cite{CHEN2020100210} developed a simulation and analysis pipeline for traffic evacuation policy to explore different aspects of evacuation including map creation, demand generation etc. These models are useful for evacuation planning before the hurricane event as they can incorporate projected hurricane scenarios, future evacuation orders, hypothetical demand, etc. However, during an unfolding hurricane event, these models may not be suitable for real-time evacuation traffic prediction as they have limited ability to quickly adapt to changes in traffic conditions, detector availability, evacuation order, etc. observed on real-time data. One of the major limitations of these approaches is that they depend on predefined behavioral assumptions which make these models sensitive to evacuation behavior, hurricane characteristics, and location attributes. A simulation-based prediction model for one hurricane may not be applicable to another hurricane without proper calibration of the model parameters related to evacuee's behavior, hurricane path, evacuation zones, and other event-specific conditions. However, the model calibration and validation process can be computationally intensive and time consuming. This may limit the practicality of applying simulation-based models during an unfolding evacuation situations, where real-time adaptation to dynamic traffic and evacuation conditions are required. As a result, the generalizability of these models across different regions or future hurricane events may be limited.

% Why data-driven deep learning model?
Data-driven approaches show some promise as they can capture complex traffic patterns directly from available data. Traditional data-driven machine learning approaches use historical traffic data to predict future traffic patterns. For example, \cite{ahn2016} presented a highway traffic prediction model based on Bayesian classifier and support vector regression (SVR) to predict traffic by finding the speed level with decreasing energy function. However, traditional machine learning approaches can hardly capture the spatial-temporal correlation of traffic variables and variation of traffic states due to the inherent high dimensionality of traffic data \citep{JIANG2022117921}. Deep learning models such as Graph Neural Network (GNN), Convolutional Neural Network (CNN), and Long Short-Term Memory (LSTM) models can effectively capture complex spatial-temporal relationships in high-dimensional traffic data \citep{RAHMAN2023104126}. As transportation networks can be represented as a graph, GNN models generally perform well in capturing the spatio-temporal variation of evacuation traffic states. Particularly, graph convolutional network (GCN) model \citep{kipf2017semi} is widely adopted due to its advantages in dealing with non-Euclidean and irregular data, such as transportation network data \citep{chen2022reinforced}. For instance, \cite{rashid2024evacuation} and \cite{RAHMAN2023104126} developed Graph Convolutional Long-Short Term Memory (GCN-LSTM) models to predict evacuation traffic flows in major highways during hurricane evacuations. Both of these studies showed that graph-based models can capture evacuation traffic states better than other traditional machine learning or deep learning approaches.

\begin{figure}[htbp]
    \centering
    \includegraphics[width=0.8\linewidth]{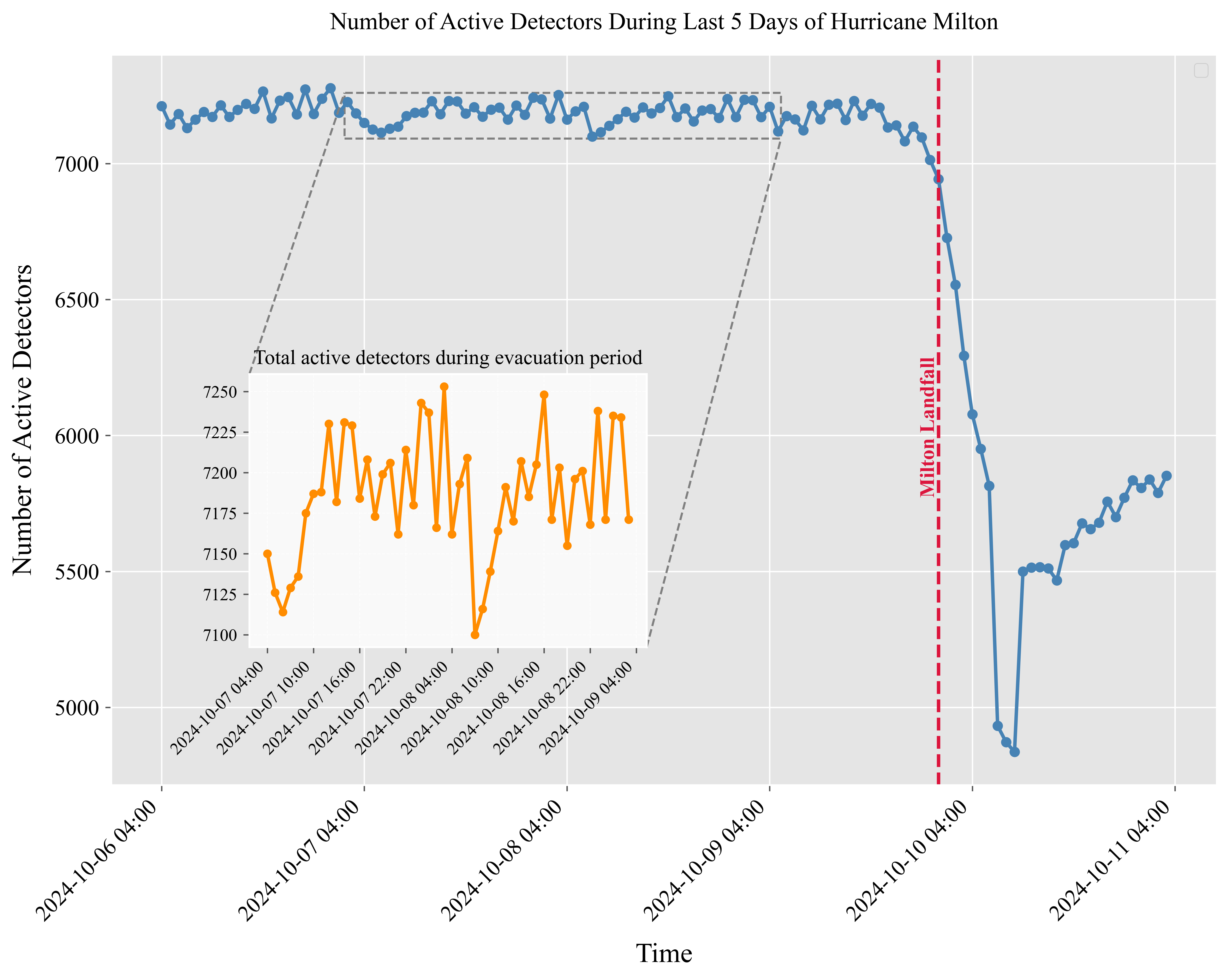}
    \caption{Network-wide total number of active detectors over time in the state of Florida, USA.}
    \label{fig:active_detectors}
\end{figure}

% Limitations of static graph-based approach and need for dynamic graph
In graph-based models, the transportation network is represented as a graph having several nodes and links. The nodes typically represent traffic detectors and the links represent the road segments connecting the detectors. Traditional GNN models consider the graph-based transportation network to be static as the number of nodes remain same over time. However, traffic detectors (graph nodes) may occasionally go offline due to scheduled maintenance, power outage or other technical reasons. Particularly during hurricane period, a significant number of detectors may go offline due to grid failure, rain, flood or other infrastructure disruptions. Figure~\ref{fig:active_detectors} shows the total number of active detectors all over Florida across different time steps during Hurricane Milton. It shows that the number of active detectors in the network is always changing with time. As the landfall approaches, the number of active detectors significantly reduces. It suggests that the graph topology is dynamic in nature, where the nodes and links are always evolving with time. Although current graph-based evacuation traffic prediction models \citep{rashid2024evacuation, RAHMAN2023104126} capture the dynamic variation in traffic features (e.g. flow or speed), they do not capture the the dynamics within the graph topology itself. Furthermore, detector-based network topology also varies across multiple hurricanes as many existing detectors may go out of order permanently and new detectors may be installed to cover unobserved locations. A graph built with only the common detectors (nodes) across all time steps may miss several detectors between a pair of active detectors. As a result, a model built using only the common detectors (static approach) may fail to accurately capture traffic flow propagation between the common detectors. Hence, the model may not generalize across the network or hurricanes, resulting in poor prediction performance for a future hurricane.

% Contribution of this work
To address this problem, we propose a novel generalized dynamic graph learning framework to capture the dynamics within the transportation network for predicting traffic during hurricane evacuation. We developed a Graph Convolutional Long-Short Term Memory (GCN-LSTM) model to predict evacuation traffic flows in five major highways (I-4 East, I-10 West, I-75 North, I-95 North and Florida's Turnpike North) of Florida. We used traffic data from twelve (12) historical hurricanes which made landfall in the state of Florida from 2016 to 2024. For each hurricane, we collected ten (10) consecutive days of traffic data including non-evacuation period, evacuation period and landfall day data. As the hurricane evacuation period typically lasts only for 1 to 5 days, we also used some non-evacuation traffic data for a better understanding of traffic patterns. Hence, the model captures the transition of traffic flows from non-evacuation to evacuation traffic scenarios. We also collected evacuation related data during each hurricane and integrated it with traffic data. After processing all hurricane data, we developed a dynamic graph construction framework that accounts for dynamic detector outages over time to capture changes in detector availability and generates a unique graph topology at each time step. Then, we trained a GCN-LSTM model for predicting evacuation traffic integrating the generated dynamic graphs. We tested the model on the two-day evacuation period of Hurricane Milton: October 7 and 8, 2024. 

In this study, we adopted a modular approach for training the model, where we provided each hurricane's data as a different module. To be specific, we first trained the model on Hurricane Helene (2024), which is the immediate past hurricane before Hurricane Milton (2024) and tested the model on Milton's two-days evacuation period. Then, we added Hurricane Debby's data to the training data, hence training the model on both Hurricanes Debby (2024) and Helene (2024) and tested it on Hurricane Milton (2024). We kept adding past hurricanes data like that in the training process in a sequential order. This modular approach helps to track the performance of our model and gives us an idea about the amount of training data requirement for optimal model performance. Following this modular approach, we finally trained the model on traffic and evacuation related data of eleven (11) historical hurricanes ranging from Hurricane Hermine (2016) to Hurricane Helene (2024) and predict Milton's evacuation traffic flow with a lead time up to 6 hours. To test the generalization capability of the model, we also trained another model on past hurricanes from Hurricane Hermine (2016) to Hurricane Elsa (2021) and tested on Hurricane Ian (2022). 

The framework is designed to generalize across hurricanes in Florida without requiring retraining or fine-tuning for each new hurricane event. In practice, this means that as soon as a new hurricane forms and its projected path and probable landfall time information become available, emergency managers can generate reliable, network-wide real-time traffic forecasts using the proposed framework and take interventions in highly congested regions. This is particularly useful during rapidly intensifying hurricanes (RIH), when agencies may have limited time to prepare \citep{Bhatia2019_RI,Lipiec2024_RI}. By avoiding the need for hurricane-specific retraining or finetuning, the proposed framework can provide traffic forecasts more readily and support timely evacuation management decisions.

% Contribution: bullet points
In summary, we present a novel methodological framework that can capture the dynamic spatial-temporal dependencies of evacuation traffic states and predict future traffic flows considering the evolving network topology across multiple hurricanes. The contributions of this study are summarized as follows:

\begin{enumerate}
    \item We develop a novel dynamic graph construction framework that accounts for dynamic detector outages over time to represent evolving detector-based transportation network structure under different hurricane conditions. Previous graph learning models mostly relied upon a fixed graph network structure.  
    \item We introduce a modular training approach to show the incremental gain in predictive performance of dynamic graph-based traffic prediction models over multiple hurricanes.
    \item We compile traffic and evacuation data from 12 past hurricanes that made landfall in Florida, USA and rigorously evaluate the model's generalizability---an aspect that has been overlooked in previous studies. 
    \item Considering all major freeways in Florida, we develop a network-wide generalized Graph Convolutional-Long Short Term Memory model that is trained over data from multiple past hurricanes of Florida. The model is capable to provide reliable real-time traffic predictions during the evacuations of a future hurricane in Florida without requiring target-event fine tuning.
\end{enumerate}

% Paper structure
The remainder of this paper is organized as follows. Section \ref{sec:literature} reviews relevant literature on data-driven traffic prediction models, especially focusing on evacuation traffic problems. Section \ref{sec:data} describes the study area and the data used in this study. Section \ref{sec:methodology} presents the methodology and modeling framework developed to predict evacuation traffic flows. Section \ref{sec:results} discusses the major outcomes of the model and analyzes the results. Section \ref{sec:baseline_and_ablation} conducts model ablation and compares the proposed model with other baseline models. Section \ref{sec:conclusions} concludes the paper with key findings, limitations, and potential future research directions.

\section{Literature Review} \label{sec:literature}
% Evacuation behavior studies and need for traffic detector data
Efficient evacuation traffic management relies on how accurately we can predict future traffic condition in real time \citep{RAHMAN2023104126}. Existing literature on evacuation traffic management mostly focuses on statistical methods and behavioral analysis of evacuees to understand evacuation traffic pattern during major emergency events such as hurricane, wildfire, flood etc. For example, \cite{lindell2018large} discussed how individuals respond towards evacuation orders by examining the factors influencing evacuation compliance, timing, and destination choices. \cite{huang2016who} conducted a statistical meta-analysis of 49 studies on hurricane evacuation behavior and identified key factors that influence individual decisions to evacuate, such as official warnings, residence in mobile homes, and perceived personal risk. They noted that demographic variables generally have weaker effects on evacuation decision-making. A major limitation of these behavioral studies is that they require survey data, which is expensive and difficult to collect during evacuation period. Using real-time traffic detector data can shed some light in this regard to predict evacuation traffic patterns for better management.

% Evacuation decision support tool and need for a generalized model
Predicting evacuation traffic in advance is challenging due to uncertainty of hurricane intensity and landfall location, leading to uncertain evacuation decision making. Many studies focused on individual- or household-level decision-making models to anticipate evacuation traffic. For instance, \cite{yang2019hurricane} developed an integrated modeling framework designed to enhance hurricane evacuation decision-making under uncertainty by combining hazard modeling, behavioral analysis, and traffic simulations. \cite{blanton2020integrated} presented a physics-based hazard modeling approach that dynamically accounts for the physical interactions among hazard components and captures hurricane evolution uncertainty using an ensemble method. To improve evacuation planning, \cite{davidson2020integrated} introduced a framework that integrates probabilistic hazard scenarios, dynamic population behavior models, and traffic simulations within a multistage stochastic programming approach. However, these decision support models mostly rely on predefined scenarios and may not adapt swiftly to real-time changes in hurricane characteristics or evacuation dynamics. They also make assumptions on human decision making that may not accurately reflect the complexities and heterogeneity in evacuee responses. These limitations warrant the need for a data-driven generalized evacuation traffic prediction model using real-time traffic detector data.

% Existing data-driven models and their limitations
Several recent studies focused on developing data-driven models for predicting evacuation traffic. For example, \cite{reza2018speed} developed a data-driven short-term traffic speed prediction model during hurricane evacuation using Long-Short Term Memory (LSTM) neural network. \cite{RAHMAN2023104126} developed a deep learning model adopting Graph Convolutional Network (GCN) and LSTM framework for predicting evacuation traffic during Hurricane Irma using traffic detector data. Recently, \cite{rashid2024evacuation} developed a GCN-LSTM model for evacuation traffic prediction during Hurricane Ian by using traffic detector and Facebook movement data. One of the major limitations of these studies is that these models were primarily trained using non-evacuation period traffic data, which may fail to capture diverse evacuation traffic scenarios. Additionally, all the models discussed above were hurricane-specific, i.e., they only focus on individual hurricane. Hence, they may not be generalized for predicting evacuation traffic for future hurricanes. Moreover, both \cite{RAHMAN2023104126} and \cite{rashid2024evacuation} adopted a transfer learning approach, which exposed some part of the evacuation related information of the current hurricane to the trained model for fine-tuning model parameters for evacuation scenario. However, evacuation information may not be available beforehand for a future hurricane. Also, it is not practical to re-train a model with the data from an unfolding evacuation scenario. Hence, those models may face implementation challenges for predicting evacuation traffic in real-time. Another major limitation of the graph-based approaches discussed above is that they considered a static graph topology during the training process representing a static detector-based transportation network. However, Figure~\ref{fig:active_detectors} shows that the structure of detector based transportation network is highly dynamic in nature, evolving to new network structure at every time step. As static graph framework fails to capture the dynamics within the transportation network, a dynamic graph representation can help in this context to better understand evolving evacuation traffic scenarios.

% Literature on dynamic graph
Dynamic graph representation has been studied in some recent literature, showing the effectiveness of modeling the dynamic structures of graphs for improving performance in general node classification and link prediction tasks. For instance, \cite{pareja2020evolvegcn} proposed a dynamic graph learning approach by evolving the parameters of GCN model over time using recurrent neural networks (RNNs), enabling the adaptation to node and edge changes without requiring fixed node embeddings. \cite{Cui2024dynamic} proposed a dynamic extension of GCN model which updates the node embeddings over time to enable efficient graph representation learning. \cite{gan2024multigraph} proposed a multi-graph fusion framework for dynamic graph convolutional network that jointly learns graph structures and low-dimensional feature representations by integrating local and global graph information to improve classification accuracy. However, none of the above studies have employed dynamic graphs to represent the evolving detector-based transportation network.

% Dynamic graph for traffic prediction
Some recent studies have used dynamic graphs for traffic prediction, where the graph structure was allowed to change over time. For example, \cite{peng2021dynamic} proposed a dynamic graph convolutional network that integrates reinforcement learning (RL) to generate adaptive traffic graphs and combines them with LSTM for long-term traffic flow prediction under data defects. \cite{zhang2021multiple} introduced evolution temporal graph convolutional network (ETGCN), a dynamic graph-based model that fuses multiple spatial correlation matrices and evolves GCN parameters with Gated Recurrent Units (GRUs) to predict traffic speed over time. \cite{chen2022reinforced} proposed a dynamic graph convolutional network model for traffic flow prediction using deep reinforcement learning and multi-graph fusion, where the graph adjacency matrix is adaptively generated to represent the spatio-temporal dependencies between graph nodes. However, none of these studies considered the dynamic change of the detector-based network based on detector outage. Additionally, these studies mainly focused on regular traffic conditions. Their applicability to hurricane evacuation traffic is still limited because evacuation traffic is affected by sudden demand surges, changing evacuation orders, uncertain hurricane trajectories, and detector outages.

% Our contributions
To address these gaps, we propose a generalized dynamic graph learning framework for real-time, network-wide evacuation traffic prediction under changing detector availability. The proposed framework incorporates dynamic graph construction to represent detector outages over time and evaluates model generalization across multiple hurricanes without requiring target-event fine-tuning. This framework can be generalized for future hurricanes across the transportation network of Florida and can assist in evacuation traffic management. In practice, the network-wide traffic forecasts generated by this framework can support more efficient and adaptive evacuation management under evolving traffic conditions. These forecasts can help transportation and emergency management agencies identify potential congestion hotspots in advance, monitor congestion propagation, and allocate operational resources to critical locations. The predicted traffic conditions can also support proactive decisions related to field personnel deployment, law enforcement positioning, roadside assistance, emergency shoulder use, and coordinated traffic management across jurisdictions.

\section{Study Area and Data Description} \label{sec:data}
\begin{figure}[h]
    \centering
    \includegraphics[width=0.8\linewidth]{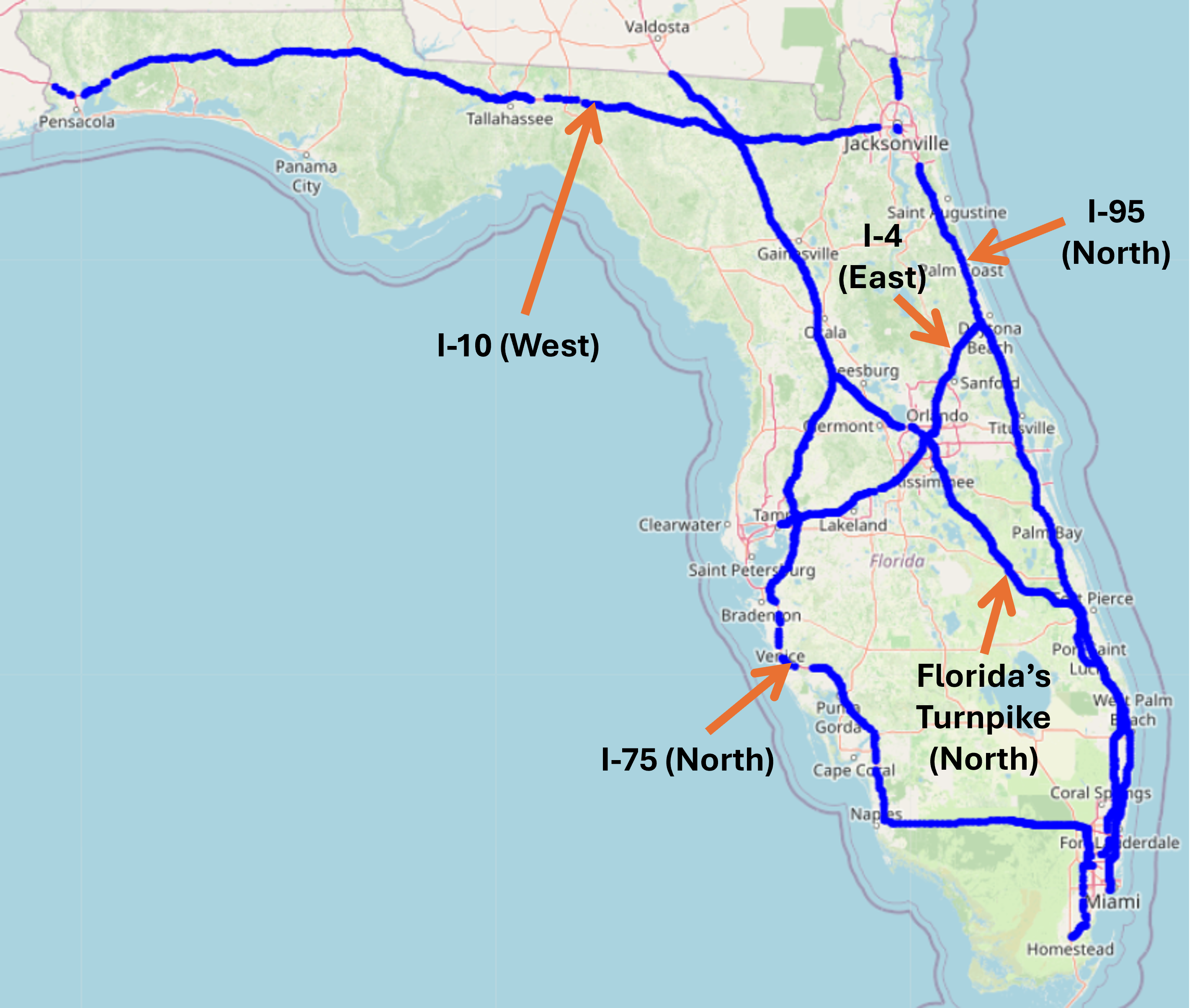}
    \caption{Florida Traffic Detector Distribution in Major Highways}
    \label{fig:det_plot}
\end{figure}

% Traffic data description
In this study, we considered a transportation network with five major interstate highways of Florida: I-4 (East), I-10 (West), I-75 (North), I-95 (North) and Florida's Turnpike (North). Figure~\ref{fig:det_plot} shows the location of the detectors in those five highways. Depending on landfall location, all or some of these highways serve as major evacuation routes during any hurricane in Florida, through which people move to Georgia or other adjacent states \citep{RAHMAN2023104126}. We collected the traffic data of these highways from Regional Integrated Transportation Information System (RITIS) \citep{ritis_archive}. RITIS collects data from Microwave Vehicle Detection System (MVDS) detectors deployed in all major roads and highways of Florida by Florida Department of Transportation (FDOT). RITIS provides detector level traffic volume, speed and occupancy data at a high resolution. For each hurricane, we collected the hourly traffic data for ten (10) days period (see Table~\ref{tab:hurricane_details}), spanning non-evacuation, evacuation and landfall day data. We process the raw detector data to discard detectors containing significant portion of missing values and zero values. After processing the data, we build a static master graph network combining all the detectors information. The details about the detector data processing and graph construction are discussed in Section~\ref{sec:methodology}.

\begin{table}[h]
\centering
\caption{Hurricane Information and RITIS Data Collection Period}
\label{tab:hurricane_details}
\begin{tabularx}{\linewidth}{|p{3.5cm}|p{5cm}|X|}
\hline
\textbf{Hurricane Name} & \textbf{Landfall Date} & \textbf{Data Range} \\
\hline
Hermine & September 2, 2016 & 08/24/2016 – 09/02/2016 \\
Matthew & October 7, 2016 & 09/28/2016 – 10/07/2016 \\
Irma & September 10, 2017 & 09/01/2017 – 09/10/2017 \\
Michael & October 10, 2018 & 10/01/2018 – 10/10/2018 \\
Dorian & September 4, 2019 & 08/26/2019 – 09/04/2019 \\
Elsa & July 7, 2021 & 06/28/2021 – 07/07/2021 \\
Ian & September 28, 2022 & 09/19/2022 – 09/28/2022 \\
Nicole & November 10, 2022 & 11/01/2022 – 11/10/2022 \\
Idalia & August 30, 2023 & 08/21/2023 – 08/30/2023 \\
Debby & August 5, 2024 & 07/27/2024 – 08/05/2024 \\
Helene & September 27, 2024 & 09/18/2024 – 09/27/2024 \\
Milton & October 10, 2024 & 10/01/2024 – 10/10/2024 \\
\hline
\end{tabularx}
\end{table}

\begin{figure}[htbp]
    \centering
    \includegraphics[width=0.8\linewidth]{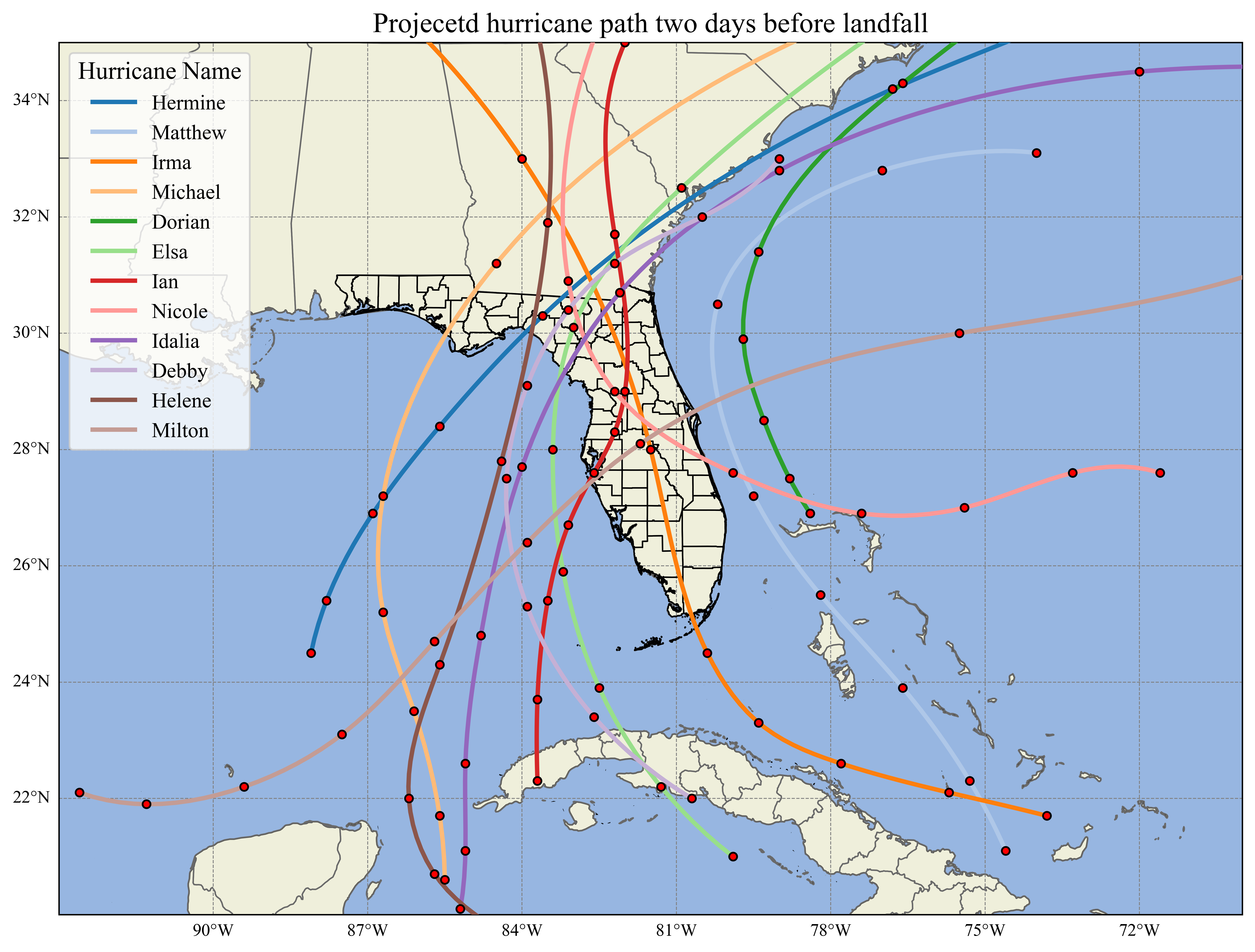}
    \caption{Projected trajectory of historical hurricanes at two days before landfall (Each point represents forecasted center of the storm along the projected hurricane trajectory. Source: \cite{nhc_archive}).}
    \label{fig:hurricane_path}
\end{figure}

% Evacuation data description
Hurricane paths are highly dynamic and uncertain in nature. We collected projected hurricane track information from the National Hurricane Center (NHC) advisory archive \citep{nhc_archive}. Figure~\ref{fig:hurricane_path} shows the projected paths of the historical hurricanes considered in this study using the forecast information available approximately two days before landfall. For each hurricane, the forecasted center locations of the storm at future time steps were extracted from the NHC advisory data and plotted (the red points). The points were then connected to illustrate the projected hurricane path of each hurricane. Evacuation orders are greatly influenced by the projected hurricane paths, based on which emergency management authorities declare evacuation orders at different period of time before hurricane landfall. For most of the hurricanes, we collected the evacuation related information from \cite{florida_evac_orders} and \cite{Anand2024HEvOD}. For some of the recent hurricanes of 2024, evacuation order time information was not available in the above sources. We collected this information from corresponding county's official social media platforms \citep{facebook_homepage, x_homepage}, where they updated the information in real time. We collected the evacuation related information for each of the hurricanes listed in Table~\ref{tab:hurricane_details}.

\begin{figure}[htbp]
    \centering
    \begin{subfigure}{0.40\linewidth}
        \centering
        \includegraphics[width=\linewidth]{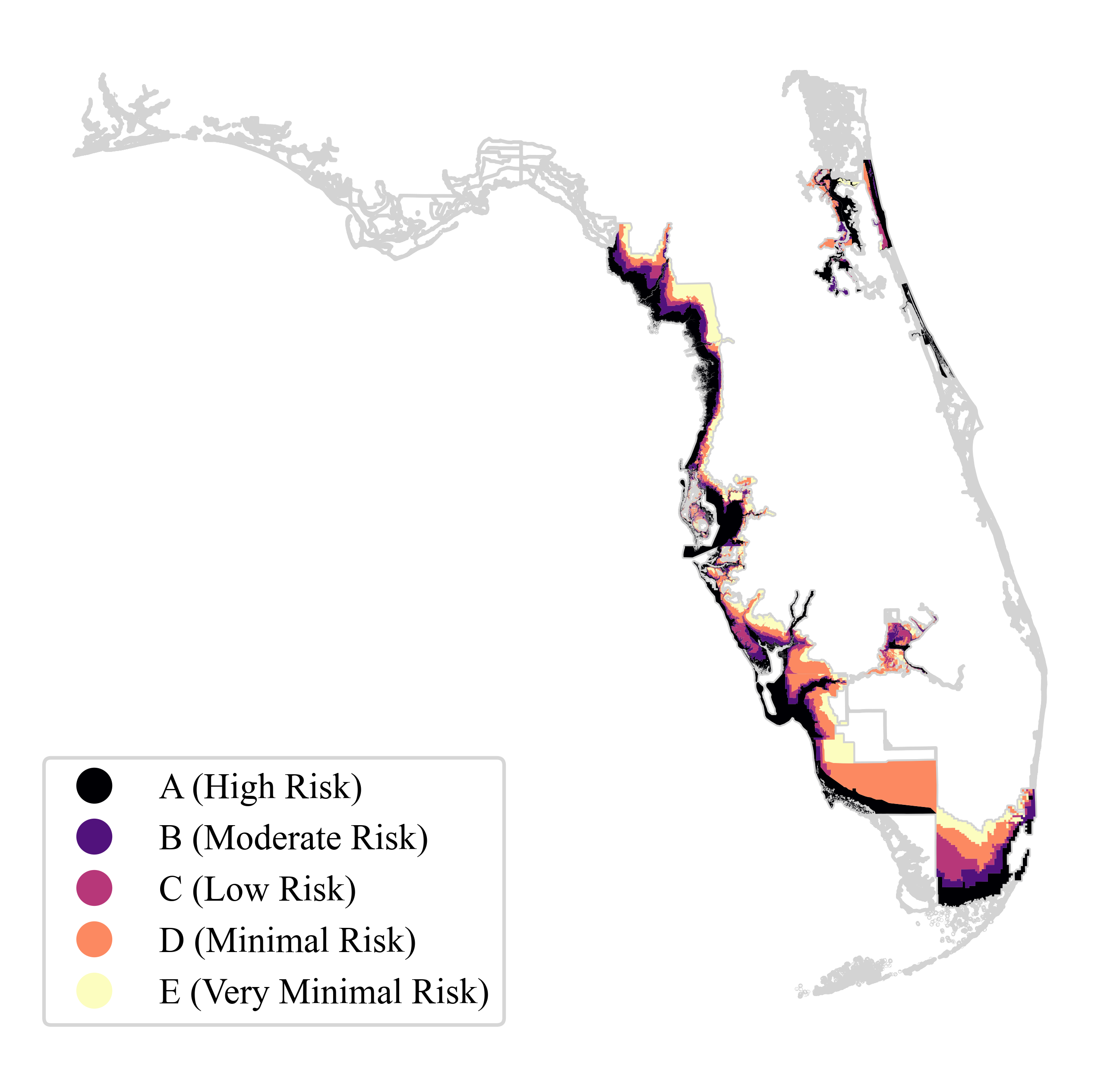}
        \caption{Evacuation zones during Milton}
        \label{fig:evc_zone_milton}
    \end{subfigure}
    \hfill
    \begin{subfigure}{0.49\linewidth}
        \centering
        \includegraphics[width=\linewidth]{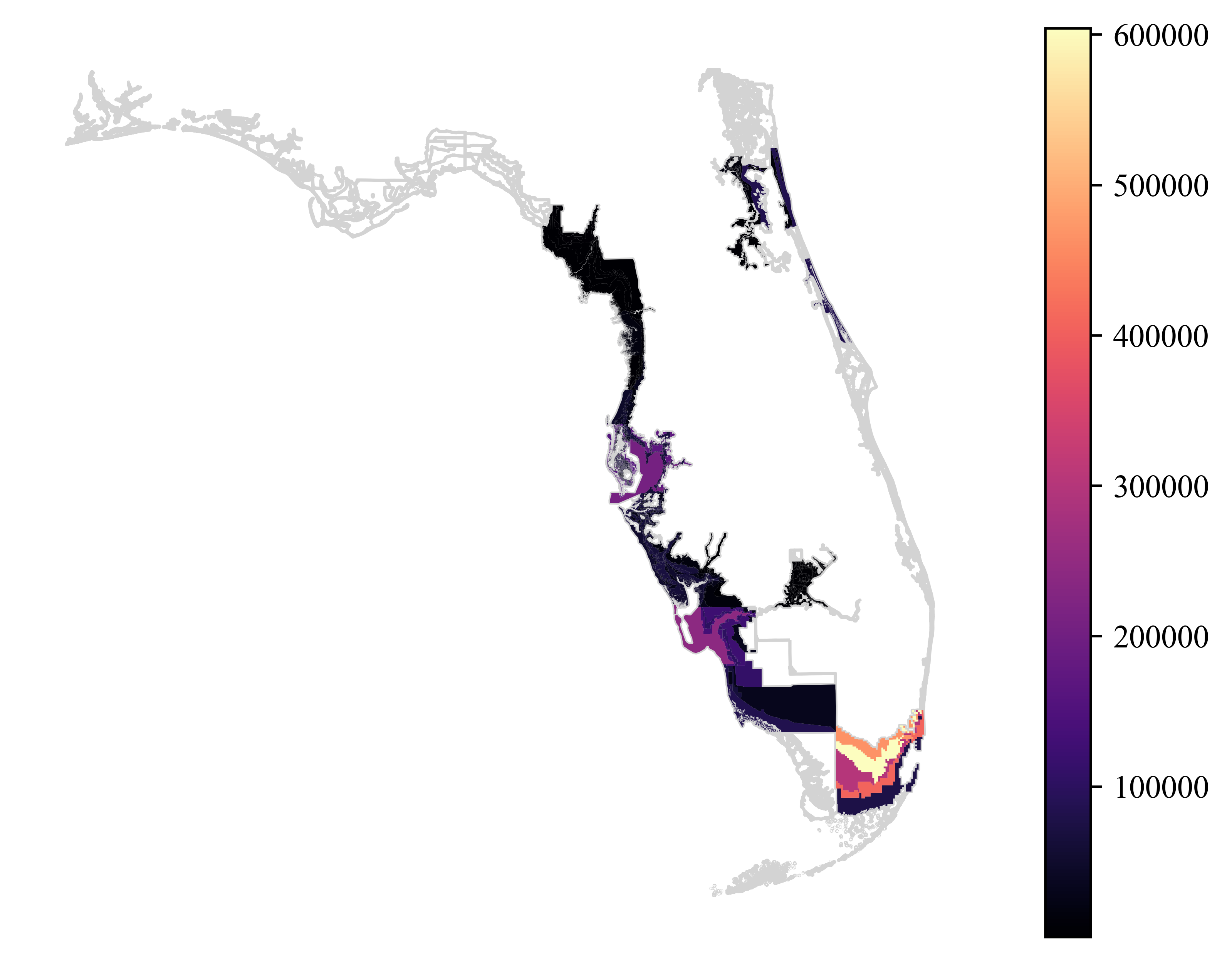}
        \caption{Population under evacuation order during Milton}
        \label{fig:evc_zone_pop_milton}
    \end{subfigure}
    \caption{Evacuation analysis during Hurricane Milton}
    \label{fig:evc_analysis_milton}
\end{figure}

Figure~\ref{fig:evc_zone_milton} shows the evacuation zones with risk profiles A to E during Hurricane Milton. The evacuation zones follow an alphabetical order where Zone A represents the highest-risk areas and Zone E represents the lowest-risk areas. These zones are designated by emergency authorities to support organized evacuation during hurricane events. Zone A generally consists of coastal areas that are more directly exposed to storm surge and hurricane impacts, while lower-risk zones are typically located farther inland \citep{RASHID2025105147}. Since Milton made landfall near the Tampa Bay area on the west coast of Florida, most of the high-risk evacuation zones were located near the west coast. Some southern and east coast regions were also under evacuation orders because of high flood risk. We also collected population data for the evacuation zones to estimate the number of people under evacuation orders during each hurricane. Figure~\ref{fig:evc_zone_pop_milton} shows the population map under evacuation order during Milton. Cumulatively, most of the people evacuates from the west coast area, causing increased traffic flow in nearby highways (I-75 North, I-4 East) compared to the regular non-evacuation period.

\section{Methodology} \label{sec:methodology}
\subsection{Data Processing and Input Features} \label{subsec:data_prep}
% Need for data processing
Raw detector data are subjected to errors due to detector malfunctioning, inclement weather, missing entries, duplicate entries and so on. Moreover, evacuation traffic face congested stop and go condition, because of which the MVDS detectors fail to detect vehicles sometimes \citep{RAHMAN2023104126, rashid2024evacuation}. For these reasons, using raw detector data might raise several complications during model training. Hence, we undertook extensive data cleaning for creating the final data for our model inputs.

\begin{figure}[htbp]
    \centering
    \includegraphics[width=0.9\linewidth]{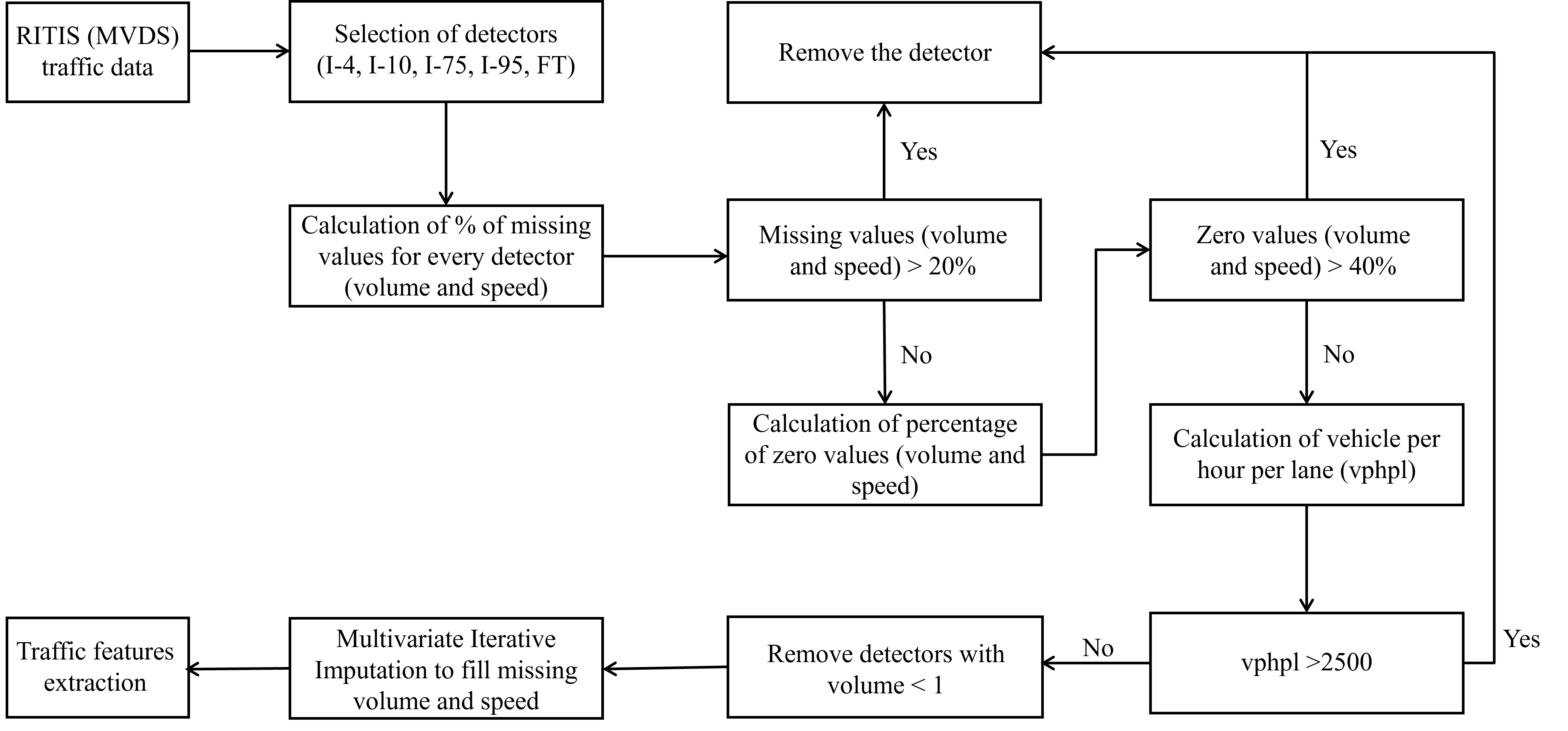}
    \caption{Traffic data processing of each hurricane}
    \label{fig:traffic_data_prep}
\end{figure}

We process each hurricane's traffic data separately. The traffic data processing steps are shown in Figure~\ref{fig:traffic_data_prep}. First, we separate the detectors located on the five major highways considered in this study from the raw RITIS traffic dataset. The percentage of missing values was calculated separately for traffic volume and speed for each detector. A detector was retained only if both traffic volume and speed contained no more than 20\% missing values. Similarly, the percentage of zero entries was calculated separately for traffic volume and speed, and detectors with no more than 40\% zero values in either variable were retained. 

For most of the hurricanes of our study, emergency authorities and FDOT allowed emergency shoulder use (ESU) to aid evacuation traffic flows. For example, FDOT authorized to use the left shoulder of I-75 northbound and I-4 eastbound to facilitate the evacuation traffic of Hurricane Milton \citep{fdot_esu}. Generally, highway capacity under regular conditions varies between 2000–2400 vehicle per hour per lane (vphpl), depending on traffic conditions and operating speeds. Operational capacity can be lower (around 2000 vphpl) during evacuation due to reduced speed \citep{reza2021evc}. However, \cite{rakesh2020} reported that ESU can increase capacity by approximately 25\%. Therefore, we considered the maximum capacity of the highways as 2500 vphpl. Detectors with observations exceeding 2500 vphpl were considered likely to be erroneous measurements and hence, those detectors were removed. We also removed the detectors having flow less than 1 vehicle per hour, as they may imply extremely sparse traffic and does not reflect the actual evacuation traffic condition.

We adopted a multivariate iterative imputation technique \citep{impute2011} using Python's scikit learn \citep{pedregosa2011scikit} library to fill the missing (NaN) volumes and speeds for existing detectors. The imputation was performed separately for each detector using traffic volume, speed, and hour of the day as input variables. We used Bayesian Ridge regression as the estimator within the iterative imputation framework to predict missing values based on the relationships among the three variables. During each iteration, missing (NaN) values were sequentially updated until convergence. To preserve realistic traffic characteristics, lower and upper bounds were determined from the observed values of each detector, and the imputed values were constrained within these ranges. It should be noted that imputation was applied only when a detector had an existing record at a given time step with missing volume or speed values. If a detector did not have any record at a given time step, no new record was created through imputation, and that detector was treated as inactive at that time step.

In order to capture the periodicity of traffic pattern, we divide 24 hours of day into 6 different time periods: Late Night (12 AM - 4 AM), Early Morning (4 AM -8 AM), Morning (8 AM - 12 PM), Noon (12 PM - 4 PM), Evening (4 PM - 8 PM), and Night (8 PM - 12 AM). We represent these features by doing one-hot encoding (0 or 1) of the time features. We also used mean and standard deviation of traffic flow in the previous day and the mean and standard deviation of traffic flow in the previous period as input features of our model to capture the flow propagation pattern over consecutive days or periods. A list of the traffic features used in the model is given in Table~\ref{tab:features}. 

\begin{table}[htbp]
\centering
\caption{Model input features}
\label{tab:features}
\begin{tabularx}{\linewidth}{X X}
\toprule
\textbf{Traffic Features} & \textbf{Evacuation Features} \\
\midrule
Detector ID & Cum. population under evacuation orders \\
Time periods & Distance from nearest evacuation zones \\
Traffic flow & Distance from landfall location \\
Traffic speed & Time left before landfall \\
Previous day mean traffic flow & Time elapsed after evacuation order \\
Previous period mean traffic flow &  Evacuation day\\
Previous day std. of traffic flow &  \\
Previous period std. of traffic flow &  \\
\bottomrule
\end{tabularx}
\end{table}

We also processed evacuation-related information gathered for each hurricane and linked this information to each detector using geo-spatial mapping. The collected evacuation information includes evacuation zones, evacuation order time, landfall time, landfall location, and population under evacuation orders. Several evacuation-related features were extracted from this information (see Table~\ref{tab:features}).

From the time-related information, we generated three features: `Evacuation day', `Time left before landfall', and `Time elapsed after evacuation order'. `Evacuation day' indicates whether a given day belongs to an evacuation period. `Time left before landfall' represents the remaining time between the current observation and hurricane landfall. `Time elapsed after evacuation order' represents the duration between the current time and the time at which the evacuation order was issued.

We also considered `cumulative population under evacuation order' to represent the number of people expected to contribute to evacuation traffic demand. However, evacuation orders do not immediately influence the traffic flow because people often delay departure decisions \citep{reza2021evc}. Previous studies showed an approximate time lag of 18 hours between evacuation order issuance and the beginning of large-scale evacuation movements \citep{reza2021evc}. Therefore, the cumulative evacuation population feature was shifted by 18 hours to account for this delayed response behavior.

Additionally, we considered distance-based evacuation features. `Distance from nearest evacuation zones' represents the shortest distance between a detector and the nearest evacuation zone. Detectors located closer to evacuation zones are expected to experience stronger evacuation-related traffic effects. `Distance from landfall location' represents the distance between a detector and the projected hurricane landfall location, which helps to capture the influence of hurricane proximity on evacuation traffic patterns.

%original
\subsection{Dynamic Graph Construction} \label{subsec:dgc}
To implement the proposed method, we constructed a detector-based transportation graph using all available detectors from the five major freeways of Florida. In the graph, each detector represents a node and the roadway segment between two consecutive detectors represents a directed edge. Two detectors are connected by a directed edge only if they are physically adjacent along the same roadway and travel direction. The direction of the edge follows the direction of the traffic . The graph is represented as \( G = (V, E, W) \), where \(V\) denotes the set of all detector nodes, \(E\) denotes the set of directed edges, and \(W\) denotes the set of edge weights.  We use the distances between nodes as edge weights in our network. We consider an weighted adjacency matrix \(A\), which represents the connectivity of the graph network, indicating the connection between nodes, the weights of which are the distances between nodes. \(A\) can be represented as follows,

\begin{equation}
A[i,j] = 
\begin{cases} 
W(i,j), & (i,j) \in E \\ 
0, & \text{otherwise}.
\end{cases}
\label{eq:adj}
\end{equation}

where \(W(i,j)\) is the edge weight between node \(i\) and \(j\), representing the distance between them. This network representation based on all possible nodes serve as a static master graph in our model. We call it a static master graph because it includes all the possible nodes of the network. Also, the network topology as well as the edge weights are not changing with time. This static master graph serves as the base for constructing the dynamic graph of our model. Figure~\ref{fig:det_plot} illustrates the detector network used to construct the static master graph. The master graph contains 2412 detector nodes and 2417 edges across the five major highways of Florida. 

During Hurricane Milton, detector availability changed considerably across these five highways. Detector inactivity ranged from approximately 32.6\% to 34.5\% of all detector-time observations (i.e., the availability of each detector at each hourly time step) from October 6 to October 9, 2024, and increased to 51.6\% on October 10, 2024. These observations suggest that a significant number of detectors may become unavailable during hurricane events, resulting in continuously evolving node availability and network topology over time. 

To account for this evolving network structure, we propose a Dynamic Graph Construction (DGC) approach for generating graph using only the active detectors at each time step. In the DGC mechanism, we first load the static master graph \(G\) of our network. At each time step \(t \in T\), we determine the active node set \(V_t \subseteq V\) by filtering the time-specific detector records from the data. A node \(v \in V\) is considered active at time \(t\) if it has a detector record at that time step \(t\) after preprocessing. The corresponding traffic flow and speed values may be original observations or imputed values for missing (NaN) observations. If a detector does not have a record at a given time step, it is treated as inactive at that time step and is not included in \(V_t\).

Considering \(V_t\), we adopt edge-based source-target filtering to generate the dynamic graphs. Specifically, for each time step \(t \in T\) and each edge \((u,v)\in E\) in the master graph \(G\), we check the availability of the source node \(u\) and target node \(v\) in the active node set \(V_t \subseteq V\). There can be four possible scenarios in this approach:

\textbf{1. Both source and target are available:} If both the source node \(u\) and the target node \(v\) are available in \( V_t\), we retain the edge \((u,v)\) and add it to the edge list \(E_t\) of the dynamic graph.

\textbf{2. Source is available, but target is not available:} If the source node \(u\) is available in \(V_t\) but the target node \(v\) is not available in \(V_t\), we conduct an Iterative Successor Search (ISS) to find the next downstream active detector along the same roadway and travel direction. The ISS procedure iteratively moves downstream from the unavailable target node \(v\) through the static master graph until it finds the first active detector \(v'\). The search follows the ordered detector sequence in the static master graph. We store the weights of the edges between the unavailable nodes from the master graph while performing ISS. We continue searching until an active node \( v' \in V_t \) is found. We build an edge \((u,v')\) to connect node \(u\) and \(v'\) and add it to the dynamic graphs' edge list \(E_t\). We sum the weights of edges between the unavailable detectors to determine the weight \(W_t(u,v')\) of edge \((u,v')\).

Let \(u=v_0 \rightarrow v_1 \rightarrow \cdots \rightarrow v_k \rightarrow v_{k+1}=v'\) be the ordered detector path from \(u\) to \(v'\), where \(v_1,\ldots,v_k \notin V_t\) are inactive intermediate detectors. Here, \(k\) denotes the number of inactive intermediate detectors between the active source node \(u\) and the next downstream active detector \(v'\). Therefore, we connect node $u$ and $v'$ with a new edge \((u,v')\) and the edge weight of \((u,v')\) is as follows,
\begin{equation}
W_t(u,v') = \sum_{i=0}^{k} W(v_i, v_{i+1})
\label{eq:wt_update}
\end{equation}

\textbf{3. Source is not available, but target is available:} This case represents the terminal edges of the graph, where the source node \(u\) is not available but the target node \(v\) is available in \(V_t\). We discard the edge \((u,v)\) in this case.

\textbf{4. Both source and target are not available:} This case also represents the terminal edges of the graph, where both the source \(u\) and target \(v\) nodes are not available in \(V_t\). We also discard the edge \((u,v)\) in this case.

The mechanism discussed above is formally presented in Algorithm~\ref{alg:dgc}. At each time step \(t \in T\), we construct a unique adjacency matrix \(A_t\) reflecting the connectivity of the generated dynamic graph \(G_t\) (see Equation~\ref{eq:A_t}). Each dynamic graph \(G_t\) is derived from the the static master graph \(G\) following Algorithm~\ref{alg:dgc}. We can express the final dynamic graph as \( G_t = (V_t, E_t, W_t) \), where \( V_t \) denotes the set of nodes at time \(t\) representing traffic detectors; \( E_t\) represents the set of directed edges at time \(t\), which defines the connectivity between nodes based on the network topology; \( W_t \) denotes the set of weights of the edges at time \(t\) that encodes the distances among the nodes. We would consider this dynamic graph \(G_t\) to build the evacuation traffic prediction model.

\begin{algorithm}[H]
\caption{Dynamic Graph Construction (DGC) Block}
\label{alg:dgc}
\KwIn{Static graph \( G = (V, E, W) \), Time steps \( T = \{t_1, t_2, \ldots, t_N\} \)}
\KwOut{Dynamic graphs \(\{G_t=(V_t,E_t,W_t) \mid t \in T\}\) and dynamic adjacency matrices \( \{A_t \mid t \in T\} \)}

\BlankLine
\textbf{Load the static graph} \( G = (V, E, W) \)\;
\For{each time step \( t \in T \)}{
    Identify active nodes \( V_t \subseteq V \)\;
    Initialize an empty dynamic graph \( G_t = (V_t, E_t, W_t) \)\;
    \For{each directed edge \( (u, v) \in E \)}{
        \If{$u \in V_t$ and $v \in V_t$}{
            Retain the edge: \( E_t = E_t \cup \{ (u, v) \} \), \( W_t(u, v) = W(u, v) \)\;
        }
        \ElseIf{$u \in V_t$ and $v \notin V_t$}{
            Perform ISS to find the next downstream active detector \(v' \in V_t\)\;
            \If{\(v'\) exists}{
                Add edge \((u,v')\) to \(E_t\)\ : \( E_t = E_t \cup \{ (u, v') \} \)\;
                Compute \(W_t(u,v')\) using Equation~\ref{eq:wt_update};
            }
        }
        \ElseIf{$u \notin V_t$ and $v \in V_t$}{
            Discard the edge: \( E_t = E_t \cup \{ \varnothing \} \)\;
        }
        \ElseIf{$u \notin V_t$ and $v \notin V_t$}{
            Discard the edge: \( E_t = E_t \cup \{ \varnothing \} \)\;
        }
    }
    Construct adjacency matrix \( A_t \) for \( G_t \):\;
    \begin{equation}
      A_t[i, j] = 
        \begin{cases} 
        W_t(i, j) & (i, j) \in E_t \\ 
        0 & \text{otherwise}
        \label{eq:A_t}
        \end{cases}  
    \end{equation}
}
\KwRet{\( G_t\) and \(A_t \) for all \(t \in T \)}
\end{algorithm}

\subsection{Problem Formulation}
In transportation networks, traffic flow dynamics can be conceptualized as a process of flow propagation, where traffic propagates from origin nodes to destination nodes through a series of intermediate nodes. This propagation creates spatial dependencies among the nodes, where traffic condition of a node can impact the traffic conditions of its neighboring nodes \citep{RAHMAN2023104126}. Such spatial correlations are critical for understanding and predicting traffic behavior across the network. In particular, congestion at one node can propagate to adjacent nodes, leading to a ripple effect that affects the entire network.

We propose a generalized dynamic graph learning framework designed to effectively capture both the spatial-temporal dependencies in evacuation traffic data and changes in evolving network topology. Our objective is to model and predict traffic flow patterns across a transportation network over multiple future time steps. The model achieves this by leveraging the spatiotemporal interdependencies among the nodes in the network, where the temporal aspect captures the evolution of traffic conditions over time, and the spatial aspect accounts for the influence of traffic conditions at neighboring nodes.

To achieve this, we employ the dynamic graph representation \(G_t=(V_t,E_t,W_t)\) of the transportation network described in section~\ref{subsec:dgc}. Let, \(\mathbf{X}_t \in \mathbb{R}^{n_t \times f}\) denote the temporal feature matrix at time \(t\), where \(f\) is the number of temporal features. These features include variables such as traffic flow, speed, temporal indicators and evacuation related temporal features. Let, \(\mathbf{Z}_t \in \mathbb{R}^{n_t \times q}\) denote the spatial feature matrix at time \(t\), where \(q\) is the number of spatial features. These features include variables such as distance from the nearest evacuation zone, distance from landfall location, etc. \(n_t=|V_t|\) denotes the number of active detector nodes at time \(t\).

The combined node feature matrix is obtained by concatenating the temporal and spatial features along the feature dimension:

\begin{equation}
\mathbf{X}_t^{C} = [\mathbf{X}_t \Vert \mathbf{Z}_t],
\label{eq:feature_comb}
\end{equation}

where \(\mathbf{X}_t^{C} \in \mathbb{R}^{n_t \times (f+q)}\), and \(\Vert\) denotes feature concatenation. Both \(\mathbf{X}_t\) and \(\mathbf{Z}_t\) are defined for the same active node set \(V_t\), so their rows are aligned by detector ID before concatenation.

The objective is to learn a mapping function \(\mathcal{F}(\cdot)\) that uses the past \(l\) time steps of dynamic graphs and node features to predict traffic flow for the next \(p\) time steps. The prediction target is denoted as \(Y_{t+1:t+p}\), where the predicted values represent traffic flow in vehicles per hour. The problem is formulated as:

\begin{equation}
\mathcal{F}
\left(
\left\{
\mathbf{X}_{\tau}^{C}, G_{\tau}
\right\}_{\tau=t-l+1}^{t}
\right)
=
\hat{Y}_{t+1:t+p}.
\label{eq:problem_formulation}
\end{equation}

Here, \(l\) is the input sequence length, and \(p\) is the prediction horizon. The model starts generating prediction when it has a complete \(l\) input sequence.

To learn this mapping function \( \mathcal{F}(\cdot) \), we develop a Graph Convolutional Long-Short Term Memory model (GCN-LSTM) architecture. The GCN-LSTM model is a spatiotemporal deep learning architecture designed to capture the spatial dependencies of nodes using Graph Convolutional Networks (GCN) and the temporal dependencies using Long Short-Term Memory (LSTM) networks. The GCN-LSTM architecture was previously adopted by \cite{RAHMAN2023104126} and \cite{rashid2024evacuation}, but they did not consider the dynamic nature of the transportation network, where the graph structure and size is changing with time. The GCN-LSTM model developed in this study can adapt with the dynamics of transportation network as well as spatial-temporal dependencies of traffic flows.

\begin{figure}[htbp]
    \centering
    \includegraphics[width=1.0\linewidth]{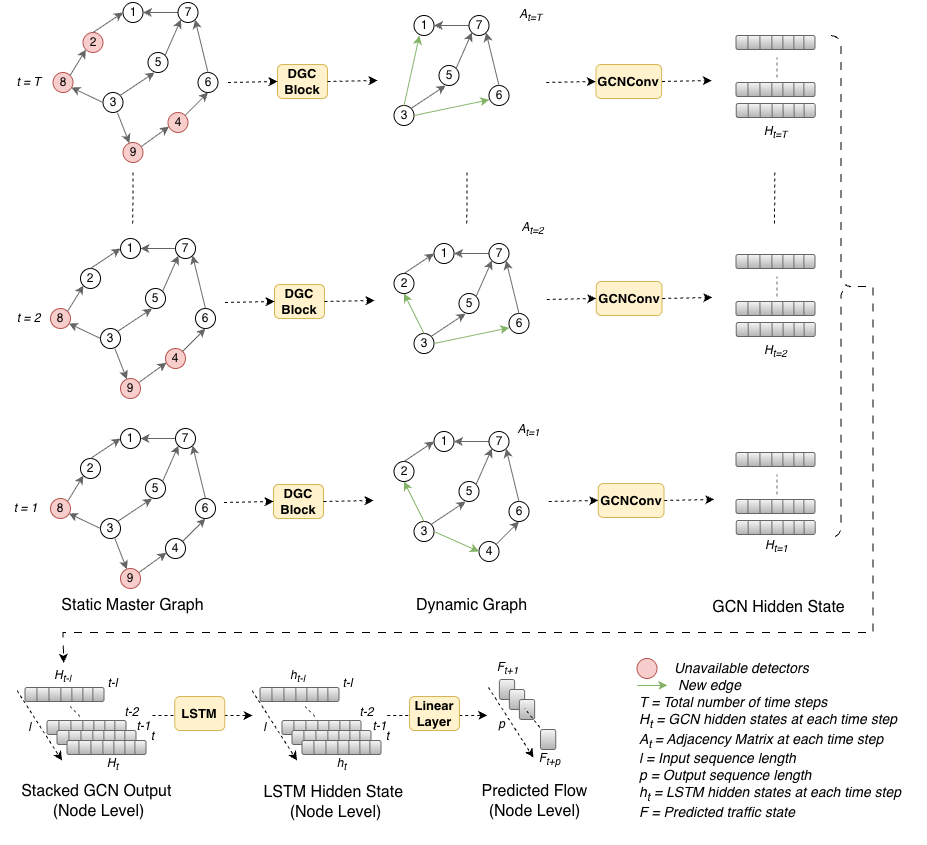}
    \caption{Generalized Dynamic Graph Learning Framework for Evacuation Traffic Prediction}
    \label{fig:dgcb}
\end{figure}

The generalized dynamic graph learning framework consists of three main components (see Figure~\ref{fig:dgcb}): the dynamic graph construction (DGC Block), the GCN layer, and the LSTM layer. The mechanism of the DGC Block is formally described in Algorithm~\ref{alg:dgc}. At each time step \( t \), the dynamic graph generated by the DGC Block is represented by an adjacency matrix \( A_t \in \mathbb{R}^{n_t \times n_t} \), where \( n_t \) is the total number of active nodes at time \(t\) (see Equation~\ref{eq:A_t}). However, directly using \( A_t \) is computationally inefficient, especially for large transportation network as most pairs of nodes are not directly connected. To overcome this, the dynamic graph \( A_t \) at each time step is transformed into an efficient sparse format rather than using a full adjacency matrix. The sparse representation only considers the connection that has non-zero edge weights and shows improved robustness and stability compared to dense representation \( A_t \) while maintaining accuracy \citep{ahmad2019dense}. The sparse representation has two components: \( \mathcal{E}_t \in \mathbb{R}^{2 \times e_t} \), known as the edge index matrix, which specifies the connections between nodes with each column defining one directed edge \( (i, j) \) such that \( A_t[i,j] > 0 \) and \( \mathcal{W}_t \in \mathbb{R}^{e_t} \), known as the edge weight vector, which stores the weights of the corresponding edges in \( \mathcal{E}_t \). Here, \(e_t=|E_t|\) denote the number of directed edges at time \(t\). The transformation is mathematically expressed as:

\begin{equation}
\mathcal{E}_t = \{ (i,j) \mid A_t[i,j] > 0 \}
\label{eq:edge_index}
\end{equation}
\begin{equation}
\mathcal{W}_t = [A_t[i,j] \mid (i,j) \in \mathcal{E}_t]
\label{eq:edge_wt}
\end{equation}

The GCN layer uses \(\mathbf{X}_t^C\), \(\mathcal{E}_t\), and \(\mathcal{W}_t\) to learn spatial dependencies among active nodes. For node \(j\), the GCN operation can be written as:
\begin{equation}
H_{t,j} = \sigma \left( \sum_{i \in \mathcal{N}_t(j) \mid (i,j) \in \mathcal{E}_t} {d_{t,i}^{-\frac{1}{2}}} {\mathcal{W}_{t,(i,j)}} {d_{t,j}^{-\frac{1}{2}}} \mathbf{X}_{t,i}^C W^{GCN} \right)
\label{eq:gcn_hidden_state}
\end{equation}

where:
 \( H_{t,j} \) is the GCN representation of node \( j \) at time \( t \);
 \( \mathcal{N}_t(j) \) is the set of neighbors of node \( j \) at time \( t \);
 \( W_{t,ij} \in \mathcal{W}_t \) is the weight of the edge between \( i \) and \( j \) at time \( t \);
 \( d_{t,i} \) and \( d_{t,j} \) are the degrees of nodes \( i \) and \( j \) at time \( t \);
 \( \mathbf{X}_{t,i}^C \) is the feature vector of neighboring node \( i \) at time \( t \);
 \( W^{GCN} \) is the trainable weight matrix of the GCN; and
 \( \sigma(\cdot) \) is the activation function (ReLU).

The GCN outputs from all time steps are then passed to the LSTM layer to model the temporal dependencies. These outputs are first stacked into a sequential form:

\begin{equation}
H_{seq} = [H_1, H_2, \ldots, H_T]
\label{eq:lstm_sequence}
\end{equation}

The LSTM processes this sequence using the following set of equations:
\begin{equation}
i_t = \sigma(W_i H_t + U_i h_{t-1} + b_i)
\label{eq:lstm_input_gate}
\end{equation}
\begin{equation}
f_t = \sigma(W_f H_t + U_f h_{t-1} + b_f)
\label{eq:lstm_forget_gate}
\end{equation}
\begin{equation}
o_t = \sigma(W_o H_t + U_o h_{t-1} + b_o)
\label{eq:lstm_output_gate}
\end{equation}
\begin{equation}
c_t = f_t \odot c_{t-1} + i_t \odot \tanh(W_c H_t + U_c h_{t-1} + b_c)
\label{eq:lstm_cell_state}
\end{equation}
\begin{equation}
h_t = o_t \odot \tanh(c_t)
\label{eq:lstm_hidden_state}
\end{equation}

where,
\(H_t\) is the GCN-generated node representation at time step \(t\);
 \( h_{t-1} \) and \( h_t \) are the hidden states at time steps \( t-1 \) and \( t \), respectively;
 \( c_t \) is the cell state at time step \( t \);
 \( i_t, f_t, o_t \) are the input, forget, and output gates of the LSTM;
 \( W_i, W_f, W_o, W_c \) are the weight matrices for the respective gates;
 \( U_i, U_f, U_o, U_c \) are the weight matrices for the recurrent connections;
 \( b_i, b_f, b_o, b_c \) are the bias terms;
 \( \sigma \) denotes the sigmoid activation function;
 \( \tanh \) represents the hyperbolic tangent activation function; and
 the operator \( \odot \) indicates an element-wise (Hadamard) product.

The final LSTM hidden state \(h_t\) (Equation~\ref{eq:lstm_hidden_state}) is used to generate the model output:

\begin{equation}
\hat{Y}_{t+1:t+p} = h_t W^{out} + b^{out}
\label{eq:output}
\end{equation}

where \(\hat{Y}_{t+1:t+p} \in \mathbb{R}^{n_t \times p}\) denotes the predicted traffic flow for the next \(p\) time steps, \(W^{out} \in \mathbb{R}^{d_h \times p}\) is the trainable output weight matrix, \(b^{out}\) is the output bias term. \(d_h\) is the LSTM hidden dimension.

The model is trained using a Mean Squared Error (MSE) loss function defined as:

\begin{equation}
L_t = \frac{1}{n_t p} \sum_{i=1}^{n_t} \sum_{h=1}^{p}
\left(Y_{t+h}^{i} - \hat{Y}_{t+h}^{i}\right)^2
\label{eq:loss_function}
\end{equation}

where, \(Y_{t+h}^{i}\) is the actual traffic flow for node \(i\) at horizon \(h\), and \(\hat{Y}_{t+h}^{i}\) is the predicted traffic flow for node \(i\) at horizon \(h\).

This GCN-LSTM based generalized dynamic graph learning architecture enables the model to effectively capture the evolving detector topology through a dynamic graph representation. It captures spatial dependencies of the dynamic graph through the GCN layer and temporal dependencies through the LSTM layer.

\subsection{Modular Training Approach}
% Training and test data
To test the performance of the proposed framework, we trained our model on historical hurricanes' traffic and evacuation related data. The dataset included hourly traffic and evacuation related features for every nodes of the dynamic graph. We used 90\% of the data for training and 10\% of the data for validation. To accommodate the varying graph sizes over time, we trained the proposed model using batch size of 1 and optimized the model parameters with Adam optimizer \citep{kingma2015adam}. We trained the model for 200 epochs with a learning rate of $1 \times 10^{-5}$. It has one GCN layer and one LSTM layer, each with 512 hidden units. To train the model, we used GPU resources provided by the Newton High-Performance Computing (HPC) cluster at the University of Central Florida \citep{ucf_newton}, allocating 164 GB of GPU memory. The training process took approximately 2 hours. Then, we tested the predictive performance of the model on the evacuation period of Hurricane Milton of October 7 and 8, 2024. We used past 6-hours of traffic and evacuation data (\(l=6\)) to predict the future evacuation traffic flow for the next 6-hour horizon (\(p=6\)).

% What is modular training and why?
A modular training approach was adopted to track the incremental learning of the model. In modular training, we did not train the model with all the training data at once. Rather, we considered each hurricane's data as a separate module in the training process. At each step, we aggregated different modules together in a sequential order so that the most recent hurricane's data stays at the end. For instance, we first trained the model with Hurricane Helene's (2024) data, which was the most recent hurricane before Milton. After training, we tested the model's performance on Hurricane Milton's evacuation period. In the next step, we added Hurricane Debby's (2024) data in the training process. Hence, we trained the model on both Hurricane Debby and Helene, and tested the performance again on Milton. We kept adding historical hurricanes data in this way in our training process and tracked the model's performance at each step. At the end, the model is trained sequentially on all eleven (11) hurricanes with Hurricane Hermine (2016) being the last of them.

% Why modular training?
Note that, we always keep the time series sequential in the modular training. At every step, the training data starts with the first time step of the most recent hurricane in that step and ends with the last time step of Hurricane Helene. This training framework helps us track the incremental gain or loss in model performance. It can also suggest when model performance starts to converge, indicating the optimal data requirement for a better performance. Further, notice that we do not use any portion of the traffic or evacuation data of Hurricane Milton at all in the modular training process. Hence, no information about future hurricane is exposed to the model beforehand. This setting is important for practice because agencies may not have enough target-event data or time to retrain a model during an unfolding hurricane. Therefore, a model without the need for target-event fine-tuning can provide a more realistic assessment of generalizability for future hurricane events.

% Check generalization capacity
Furthermore, to understand the extent of generalization capacity of our model, we tested the predictive performance on a different hurricane. Particularly, we tested it on Hurricane Ian's two-day evacuation period: September 26 and 27, 2022. This time, the training data included all the past hurricanes' traffic and evacuation related data before Ian. We also compared the performance of the models for Ian and Milton and analyzed the results.

The models were implemented using Pythons' pytorch geometric environment \citep{fey2019graph}. The performance of the model was evaluated using four metrics: Root Mean Square Error (RMSE), Mean Absolute Error (MAE), Symmetric Mean Absolute Percentage Error (SMAPE), and R\textsuperscript{2}. These metrics are defined as follows:

\begin{equation}
\text{RMSE} = 
\sqrt{\frac{1}{N} \sum_{i=1}^{N} \left(F_{\text{actual},i} - F_{\text{predicted},i}\right)^2}
\label{eq:rmse}
\end{equation}

\begin{equation}
\text{MAE} = \frac{1}{N} \sum_{i=1}^{N} \left|F_{\text{actual},i} - F_{\text{predicted},i}\right|
\label{eq:mae}
\end{equation}

% \begin{equation}
% \text{MAPE} = \frac{1}{N} \sum_{i=1}^{N} \left| \frac{F_{\text{actual},i} - F_{\text{predicted},i}}{F_{\text{actual},i}} \right| \times 100\%
% \label{eq:mape}
% \end{equation}

\begin{equation}
\text{SMAPE} = \frac{100\%}{N} \sum_{i=1}^{N} 
\frac{ \left| F_{\text{actual},i} - F_{\text{predicted},i} \right| }
     { \left| F_{\text{actual},i} \right| + \left| F_{\text{predicted},i} \right| }
\label{eq:smape}
\end{equation}

\begin{equation}
R^2 = 1 - \frac{\sum_{i=1}^{N} \left(F_{\text{actual},i} - F_{\text{predicted},i}\right)^2}
{\sum_{i=1}^{N} \left(F_{\text{actual},i} - \overline{F}_{\text{actual}}\right)^2}
\label{eq:r2}
\end{equation}

where:
 \( N \) is the total number of samples;
 \( F_{\text{actual},i} \) represents the actual flow for the \( i \)-th sample;
 \( F_{\text{predicted},i} \) represents the predicted flow for the \( i \)-th sample; and
 \( \overline{F}_{\text{actual}} \) is the mean of the actual flows.

To quantify prediction uncertainty and assess the proximity of model predictions to the actual aggregated traffic patterns, we introduced an additional metric based on the network-wide mean and standard deviation of observed hourly traffic volumes. Specifically, for each time step, we computed the standard deviation of hourly traffic volume from the network-wide aggregated ground truth. Then, we calculated the percentage of model predictions that fall within $1.5$ times the standard deviation of the corresponding observed mean. Notably, under a normal distribution assumption, approximately $87\%$ of values are expected to lie within $\pm1.5$ times the standard deviation from the mean. This metric provides a measure of how closely the predictions align with typical variations in observed traffic. 
 
\section{Results} \label{sec:results}
\begin{figure}[htbp]
    \centering
    \includegraphics[width=0.95\linewidth]{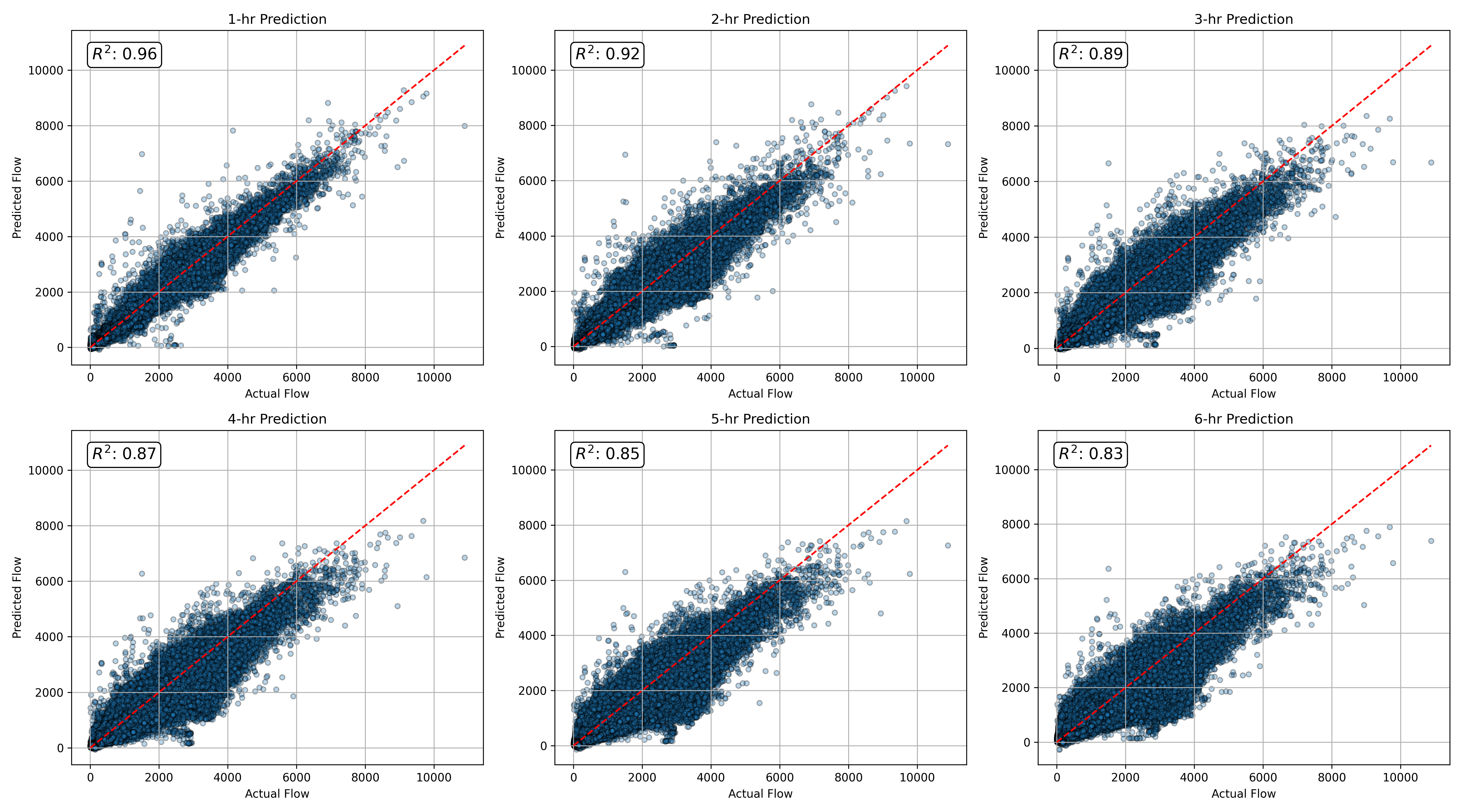}
    \caption{Comparison between actual and predicted traffic flows for 1-hour to 6-hour prediction horizons during Milton}
    \label{fig:flow_scatter}
\end{figure}

After the model formulation is complete, we tested our model for predicting the traffic of the evacuation period of Hurricane Milton. We trained the model using the data from eleven (11) historical hurricanes that made landfall before Milton and predicted traffic flows during the 2-day evacuation period of Milton. Table~\ref{tab:evacuation_metrics_milton} shows the performance of the model which achieves an RMSE value of 280.6 with an R\textsuperscript{2} of 0.96 when predicting the next-hour traffic flow. When the prediction horizon increases, the accuracy starts to drop with increasing RMSE values. It is reasonable because any predictive model shows a drop in performance if the prediction horizon is longer. Also, we have seen a gradual and stable change in performance while the prediction horizon is increased. Figure~\ref{fig:flow_scatter} shows the correlation between actual and predicted traffic flows for prediction horizons of 1 to 6 hours. This plot shows that the correlation between actual and predicted traffic flows is significantly stronger when the prediction horizon is shorter. The correlation weakens for longer prediction horizon, but this behavior is natural for predictive models where a time-series (in our case, LSTM) segment is involved. Overall, on average, the model achieves an RMSE value of 459.8 with R\textsuperscript{2} of 0.89 considering results of all prediction horizons (from 1-hour to 6-hour prediction horizons). The percentage of predictions within $1.5\sigma$ of the observed mean remains consistently high across all horizons (ranging from 92.8\% to 94.0\%), indicating that the proposed model reliably captures variations in evacuation traffic flow and offers strong predictive consistency.

% Milton 2-days
\begin{table}[htbp]
\centering
\caption{Evacuation Traffic Prediction Results for Hurricane Milton}
\label{tab:evacuation_metrics_milton}
{\scriptsize (\textit{Minimum flow = 6, Maximum flow = 10889, Mean flow = 1838, Median flow = 1503})}
\resizebox{\textwidth}{!}{%
\begin{tabular}{lccccc}
\toprule
\textbf{Prediction Horizon} & \textbf{RMSE} & \textbf{MAE} & \textbf{SMAPE} & \textbf{R\textsuperscript{2}} & \textbf{\% within 1.5$\sigma$} \\
\midrule
1-hour & 280.6 & 179.3 & 6.9 & 0.96 & 92.8 \\
2-hour & 396.5 & 256.5 & 9.3 & 0.92 & 93.4 \\
3-hour & 457.1 & 299.4 & 10.8 & 0.89 & 93.7 \\
4-hour & 497.2 & 329.2 & 12.2 & 0.87 & 94.0 \\
5-hour & 526.2 & 353.9 & 13.8 & 0.85 & 93.8 \\
6-hour & 547.3 & 370.4 & 14.4 & 0.83 & 93.9 \\
\midrule
\textbf{Overall} & \textbf{459.8} & \textbf{298.1} & \textbf{11.2} & \textbf{0.89} & \textbf{93.6} \\
\bottomrule
\end{tabular}
}
\end{table}

\begin{figure}[htbp]
    \centering
    \includegraphics[width=0.9\linewidth]{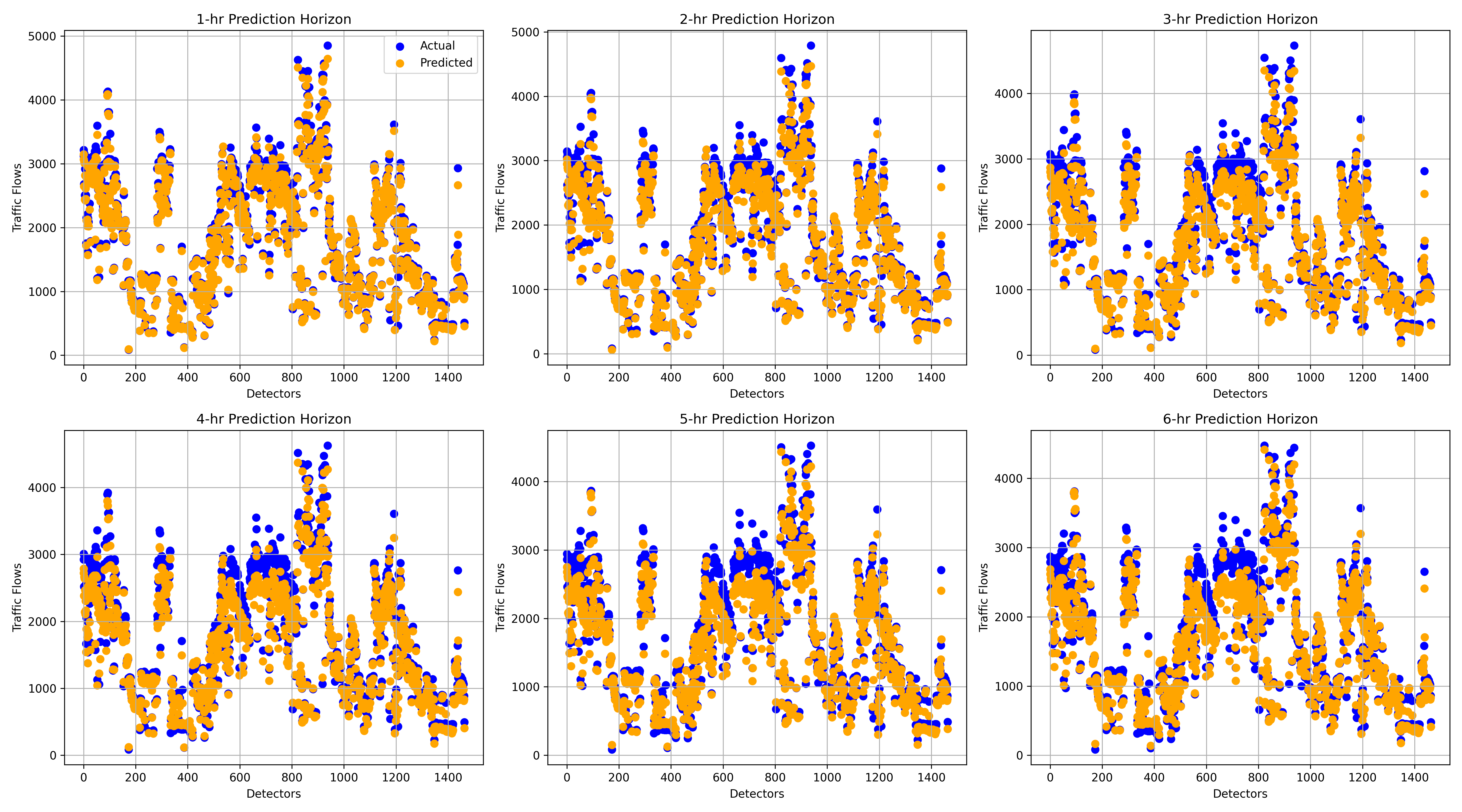}
    \caption{Detector-wise comparison of average actual and predicted flow for 1-hour to 6-hour prediction horizons during Milton}
    \label{fig:flow_dist}
\end{figure}

We also plot the detector-wise average (over the entire 2-day evacuation period) of actual and predicted traffic flows (see Figure~\ref{fig:flow_dist}) for 6-hour prediction horizon during Hurricane Milton. By comparing the average actual and predicted flows for each detector, we can assess the model's ability to generalize across different locations, each of which may exhibit unique traffic characteristics in a dynamic network. As shown by Figure~\ref{fig:flow_dist}, there is a little misalignment between the average actual and predicted flows for a small number of detectors for longer prediction horizons (visible blue dots). However, we see a close alignment across most of the detectors at shorter prediction horizons. We mentioned earlier that traffic detectors might temporarily become offline due to power outage, maintenance reason or any disruptions. As the number of active detectors at each time step is always changing, it makes the detector based transportation network very dynamic. As demonstrated in Figure~\ref{fig:flow_dist}, the model can capture the spatiotemporal dependencies of evacuation traffic flows for most of the active detectors very well in a dynamic traffic network. It suggests that, for an unseen hurricane, the model also generalizes well at the level of individual detectors when capturing the complex spatiotemporal dependencies of evacuation traffic. 

% Ian 2-days
\begin{table}[htbp]
\centering
\caption{Evacuation Traffic Prediction Results for Hurricane Ian}
\label{tab:evacuation_metrics_ian}
{\scriptsize (\textit{Minimum flow = 1, Maximum flow = 10240, Mean flow = 1561, Median flow = 1161})}
\resizebox{\textwidth}{!}{%
\begin{tabular}{lccccc}
\toprule
\textbf{Prediction Horizon} & \textbf{RMSE} & \textbf{MAE} & \textbf{SMAPE} & \textbf{R\textsuperscript{2}} & \textbf{\% within 1.5$\sigma$} \\
\midrule
1-hour & 257.9 & 159.3 & 7.2 & 0.96 & 91.7 \\
2-hour & 358.2 & 229.7 & 9.8 & 0.92 & 92.3 \\
3-hour & 413.6 & 271.4 & 11.7 & 0.90 & 92.4 \\
4-hour & 447.8 & 296.3 & 13.0 & 0.88 & 92.2 \\
5-hour & 471.3 & 315.6 & 14.1 & 0.86 & 92.5 \\
6-hour & 485.6 & 333.4 & 15.4 & 0.85 & 92.6 \\
\midrule
\textbf{Overall} & \textbf{413.2} & \textbf{267.6} & \textbf{11.9} & \textbf{0.89} & \textbf{92.2} \\
\bottomrule
\end{tabular}
}
\end{table}

To test the generalization capacity across different hurricanes, we trained another model using the proposed framework and tested it for another event of Hurricane Ian which was a rapidly intensifying hurricane making landfall in Florida in 2022. Table~\ref{tab:evacuation_metrics_ian} shows the performance of the model for Ian, achieving an overall RMSE value of 413.2 with R\textsuperscript{2} of 0.89 considering 1-hour to 6-hour prediction horizons. Compared to the results of Milton (see Table~\ref{tab:evacuation_metrics_milton}), the model achieves lower RMSE values when predicting the evacuation traffic of Ian. One plausible reason is that Hurricane Milton had higher evacuation demand than Hurricane Ian. Based on the evacuation information used in this study (cumulative population under evacuation orders), approximately 5.0 million people were under evacuation orders during Milton, compared to 2.9 million during Ian. If we compare the average traffic flows during both hurricanes, we also see that Ian's traffic flow is smaller (mean flow = 1561) compared to Milton (mean flow = 1838). For these reasons, we have seen lower RMSE values from the model for Ian. However, the overall SMAPE value for Milton remains lower than that for Ian (11.2\% in Milton vs. 11.9\% in Ian). Further, despite Milton having a higher mean flow, the percentage of predictions within $1.5\sigma$ of the observed mean is slightly higher (93.6\% in Milton vs. 92.2\% in Ian) for Milton compared to Ian, suggesting more consistent and reliable predictions under extreme flow conditions of Milton. Overall, the proposed model not only performs better for Hurricane Milton, but also generalizes very well for other hurricanes, such as Ian.

\begin{table}[htbp]
\centering
\caption{Highway-Specific Evaluation Metrics (Milton and Ian)}
\label{tab:road_specific_metrics}
\resizebox{\textwidth}{!}{%
\begin{tabular}{lcccccccccc}
\toprule
& \multicolumn{5}{c}{\textbf{I-4 (Milton)}} & \multicolumn{5}{c}{\textbf{I-4 (Ian)}} \\
\cmidrule(lr){2-6} \cmidrule(lr){7-11}
\textbf{Horizon} & \textbf{RMSE} & \textbf{MAE} & \textbf{SMAPE} & \textbf{R\textsuperscript{2}} & \textbf{\% Within 1.5$\sigma$} & \textbf{RMSE} & \textbf{MAE} & \textbf{SMAPE} & \textbf{R\textsuperscript{2}} & \textbf{\% within 1.5$\sigma$} \\
\midrule
1-hour & 299.4 & 212.3 & 4.8 & 0.95 & 85.4 & 279.0 & 190.3 & 5.4 & 0.96 & 93.4 \\
2-hour & 462.5 & 325.6 & 7.6 & 0.88 & 86.2 & 417.1 & 284.9 & 7.8 & 0.91 & 94.5 \\
3-hour & 542.7 & 387.0 & 9.4 & 0.84 & 87.2 & 477.0 & 325.8 & 9.2 & 0.88 & 94.2 \\
4-hour & 569.0 & 411.5 & 10.7 & 0.82 & 88.9 & 518.7 & 354.5 & 10.2 & 0.86 & 94.0 \\
5-hour & 599.1 & 437.5 & 11.8 & 0.81 & 90.7 & 566.0 & 392.0 & 11.6 & 0.83 & 94.6 \\
6-hour & 634.7 & 475.0 & 13.2 & 0.78 & 94.5 & 606.6 & 426.6 & 12.9 & 0.81 & 94.9 \\
\midrule
\textbf{Overall} & \textbf{529.7} & \textbf{374.8} & \textbf{9.6} & \textbf{0.85} & \textbf{88.5} & 
\textbf{489.3} & \textbf{329.0} & \textbf{9.5} & \textbf{0.88} & \textbf{94.4} \\
\midrule
& \multicolumn{5}{c}{\textbf{I-75 (Milton)}} & \multicolumn{5}{c}{\textbf{I-75 (Ian)}} \\
\cmidrule(lr){2-6} \cmidrule(lr){7-11}
1-hour & 317.5 & 211.2 & 7.0 & 0.95 & 95.8 & 278.1 & 181.2 & 7.3 & 0.95 & 93.2 \\
2-hour & 457.5 & 304.8 & 10.1 & 0.89 & 96.1 & 404.3 & 274.9 & 10.1 & 0.90 & 93.9 \\
3-hour & 531.6 & 354.8 & 11.3 & 0.84 & 96.4 & 464.6 & 321.9 & 12.1 & 0.87 & 94.2 \\
4-hour & 598.0 & 400.6 & 13.1 & 0.80 & 96.5 & 499.0 & 344.6 & 13.3 & 0.84 & 94.0 \\
5-hour & 647.0 & 439.3 & 15.6 & 0.76 & 96.2 & 532.0 & 369.6 & 14.4 & 0.82 & 94.1 \\
6-hour & 676.4 & 460.7 & 16.3 & 0.74 & 96.0 & 545.9 & 388.7 & 16.4 & 0.81 & 94.1 \\
\midrule
\textbf{Overall} & \textbf{551.7} & \textbf{361.9} & \textbf{12.3} & \textbf{0.83} & \textbf{96.2} & 
\textbf{463.1} & \textbf{313.5} & \textbf{12.3} & \textbf{0.87} & \textbf{93.9} \\
\bottomrule
\end{tabular}
}
\end{table}

Since both hurricanes Milton and Ian made landfall on the west coast of Florida, I-4 eastbound and I-75 northbound were the two major evacuation routes during those hurricanes. Hence, we analyzed the model performance for these two highways having extreme evacuation traffic conditions. Table~\ref{tab:road_specific_metrics} shows the model performance for both I-4 and I-75 during Milton and Ian. Overall, the model performance slightly degrades for both I-4 and I-75 compared to the network-wide prediction performance during Milton and Ian. During Hurricane Milton, the percentage of predictions within $1.5\sigma$ is consistently higher for I-75 than I-4 (96.2\% vs. 88.5\%), suggesting more stable predictions along I-75 northbound corridor. However, during Hurricane Ian, this gap narrows (93.9\% for I-75 vs. 94.4\% for I-4), indicating improved predictive consistency across both corridors. Additionally, if we notice the SMAPE values for both hurricanes, we find that the model demonstrated a stable performance for these two highways. To be specific, the model shows an SMAPE value of 4.8\% only when predicting the next-hour evacuation traffic on I-4 during Milton. Moreover, if we compare the SMAPE across hurricanes, the model showed nearly similar performance while predicting for I-4 and I-75 during Milton and Ian. 
Though Milton's evacuation scenarios were more intense than Ian, our proposed model was able to show a very stable performance for Milton on these two major evacuation routes of I-4 and I-75.

\begin{figure}[htbp]
    \centering
    \includegraphics[width=0.9\linewidth]{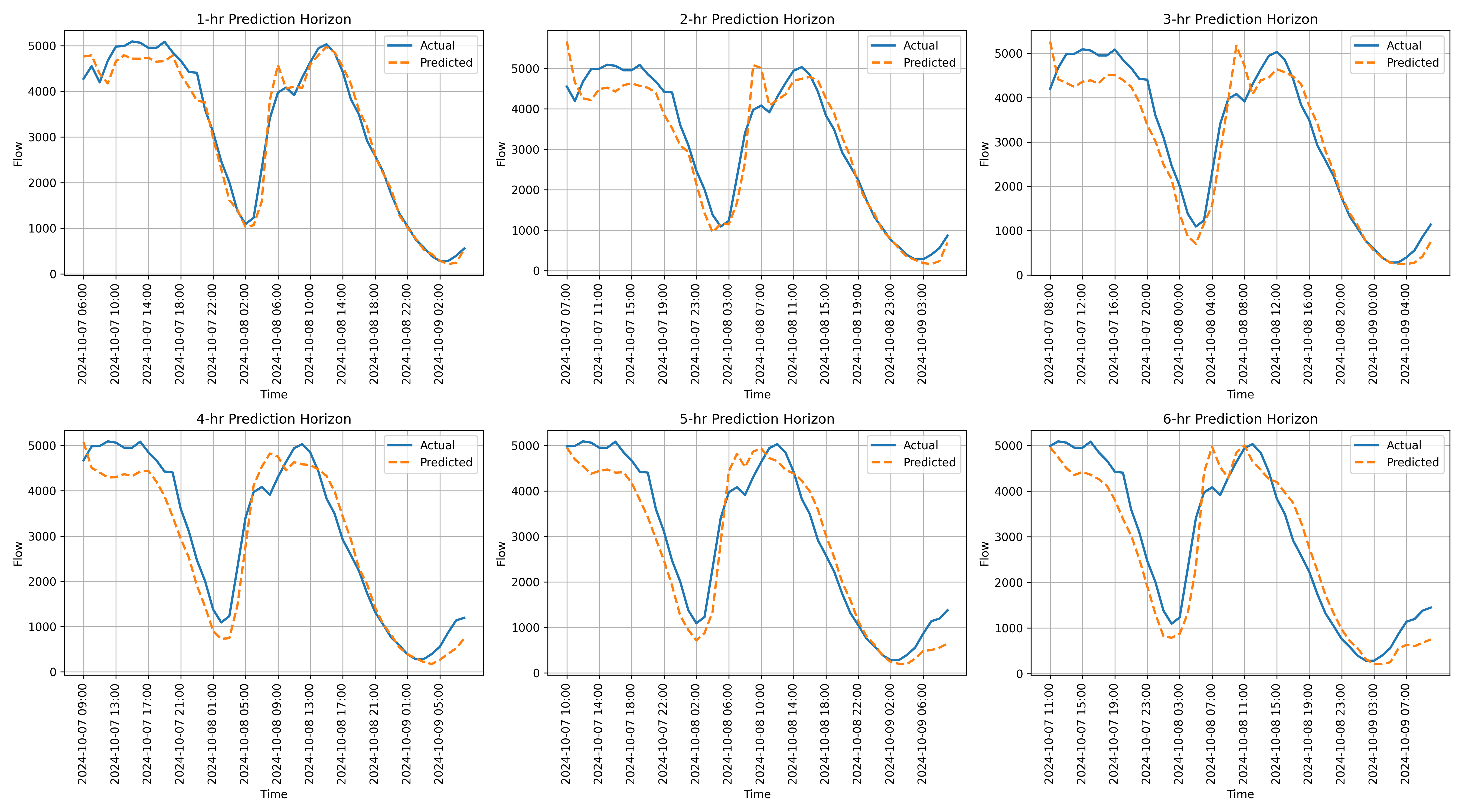}
    \caption{Model performance of an I-4 detector for 6-hour prediction horizon during Milton}
    \label{fig:flow_line_i4}
\end{figure}

\begin{figure}[htbp]
    \centering
    \includegraphics[width=0.9\linewidth]{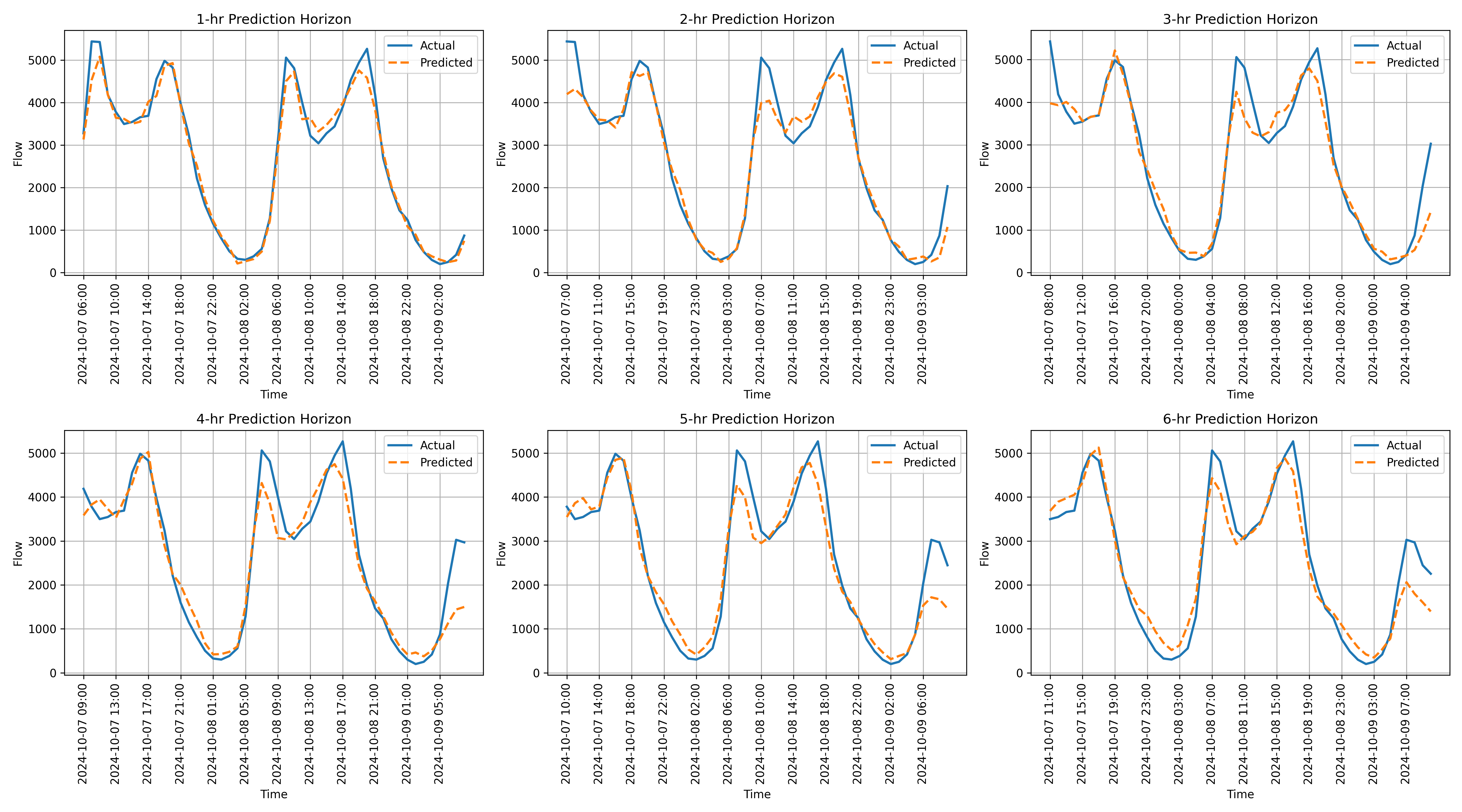}
    \caption{Model performance of an I-75 detector for 6-hour prediction horizon during Milton}
    \label{fig:flow_line_i75}
\end{figure}

We also plot the predicted and actual traffic flows of an individual detector of I-4 and I-75 for 6-hour prediction horizon during Milton (see Figure~\ref{fig:flow_line_i4} and \ref{fig:flow_line_i75}). For both detectors of I-4 and I-75, the model captures the actual flow pattern very accurately for the next-hour horizon, including the all the peaks and saddle points. For I-4, the model sometimes over-estimate the peak flow (in resemblance to the morning peak) for longer prediction horizon. On the other hand, the model sometimes under-estimate the peak flow of I-75 detector when the prediction horizon is long. However, the overall evacuation traffic pattern is captured very well for both of the detectors.

\begin{table}[!ht]
\centering
\caption{Systematic analysis of controlled upstream detector outages within input horizons for a detector at I-4}
\label{tab:input_horizon_outage_sensitivity_uid258}
\renewcommand{\arraystretch}{2.0}
\resizebox{\textwidth}{!}{%
\begin{tabular}{c|c|cc|cc|cc}
\hline
\multirow{2}{*}{\makecell{\textbf{No. of Immediate}\\\textbf{Upstream Detector Outage}}} 
& \multirow{2}{*}{\makecell{\textbf{Baseline}\\\textbf{SMAPE (\%)}}}
& \multicolumn{2}{c|}{\makecell{\textbf{33\% Outage}\\\textbf{(2 input hours outage)}}} 
& \multicolumn{2}{c|}{\makecell{\textbf{67\% Outage}\\\textbf{(4 input hours outage)}}} 
& \multicolumn{2}{c}{\makecell{\textbf{100\% Outage}\\\textbf{(all 6 input hours outage)}}} \\
\cline{3-8}
& 
& \makecell{\textbf{Outage}\\\textbf{SMAPE (\%)}} 
& \makecell{\textbf{\%}\textbf{\(\Delta\)}} 
& \makecell{\textbf{Outage}\\\textbf{SMAPE (\%)}} 
& \makecell{\textbf{\%}\textbf{\(\Delta\)}} 
& \makecell{\textbf{Outage}\\\textbf{SMAPE (\%)}} 
& \makecell{\textbf{\%}\textbf{\(\Delta\)}} \\
\hline

1 upstream 
& \multirow{5}{*}{18.83} 
& 18.84 & 0.05
& 18.84 & 0.08
& 18.85 & 0.13 \\
\cline{1-1}\cline{3-8}

2 upstream 
&  
& 18.86 & 0.19
& 18.88 & 0.29
& 18.93 & 0.57 \\
\cline{1-1}\cline{3-8}

3 upstream 
&  
& 18.89 & 0.32
& 18.92 & 0.48
& 18.99 & 0.86 \\
\cline{1-1}\cline{3-8}

4 upstream 
&  
& 18.93 & 0.54
& 18.98 & 0.84
& 19.10 & 1.44 \\
\cline{1-1}\cline{3-8}

5 upstream 
&  
& 18.98 & 0.81
& 19.03 & 1.09
& 19.14 & 1.64 \\
\hline
\end{tabular}%
}
\renewcommand{\arraystretch}{1.0}
\end{table}

To examine how upstream detector outages affect downstream detector's prediction performance, we conducted a systematic analysis focusing on a randomly selected detector on the I-4 evacuation corridor for 2-days evacuation period of Milton. We introduced artificial outages of the immediate upstream detectors of that single detector within the 6-hour input horizon used by the model. Specifically, we masked the upstream detector information for 2 input hours, 4 input hours, and all 6 input hours, to represent 33\%, 67\%, and 100\% detector outage scenarios, respectively. We also varied the number of immediate upstream detectors affected by the outage from 1 to 5. For each upstream detector outage, the network was reconstructed following Algorithm~\ref{alg:dgc}. Table~\ref{tab:input_horizon_outage_sensitivity_uid258} shows that the prediction error increases gradually as both the number of upstream detectors with outages and the input-horizon outage rate increase. The baseline SMAPE for the target detector was 18.83\%. When only one upstream detector was affected, the SMAPE increase was very small, ranging from 0.05\% to 0.13\%. However, as more upstream detectors were affected, the degradation became more noticeable. The largest increase occurred when five immediate upstream detectors had outages for the full 6-hour input horizon, where SMAPE increased from 18.83\% to 19.14\%, corresponding to a 1.64\% relative increase. The results show a consistent performance degradation pattern, although the magnitude of degradation is very small. This indicates that the prediction performance of the downstream detector is affected (by a small margin) when multiple consecutive upstream detectors are unavailable for a larger portion of the input horizon. However, the model remains relatively robust to upstream detector outages at varying outage rates for the tested case. These findings indicate that the proposed dynamic graph construction does not create severe distortion in traffic propagation.

\begin{figure}[htbp]
    \centering
    \begin{subfigure}{1.0\linewidth}
        \centering
        \includegraphics[width=\linewidth]{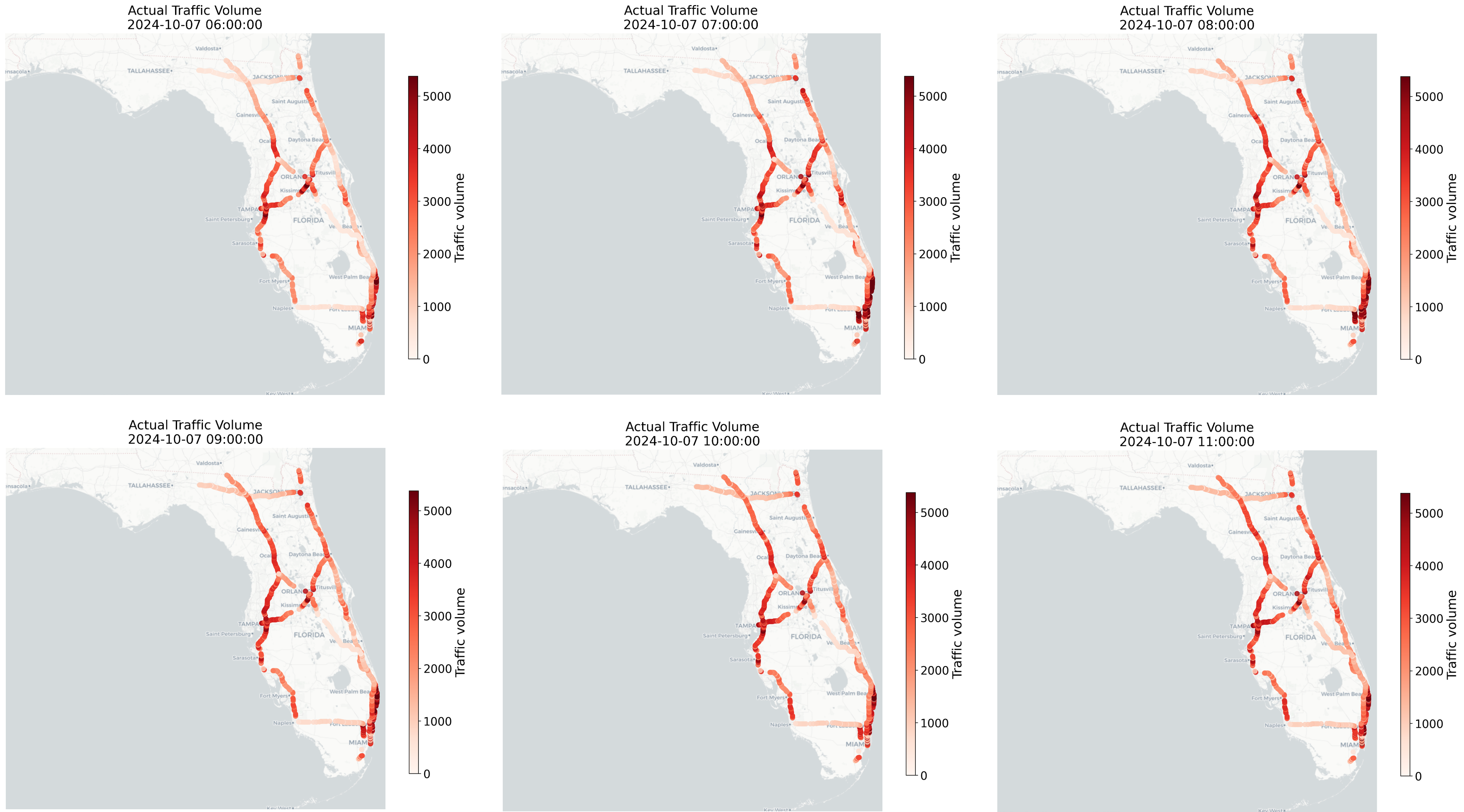}
        \caption{Congestion propagation of actual traffic}
        \label{fig:con_actual}
    \end{subfigure}
    \hfill
    \begin{subfigure}{0.99\linewidth}
        \centering
        \includegraphics[width=\linewidth]{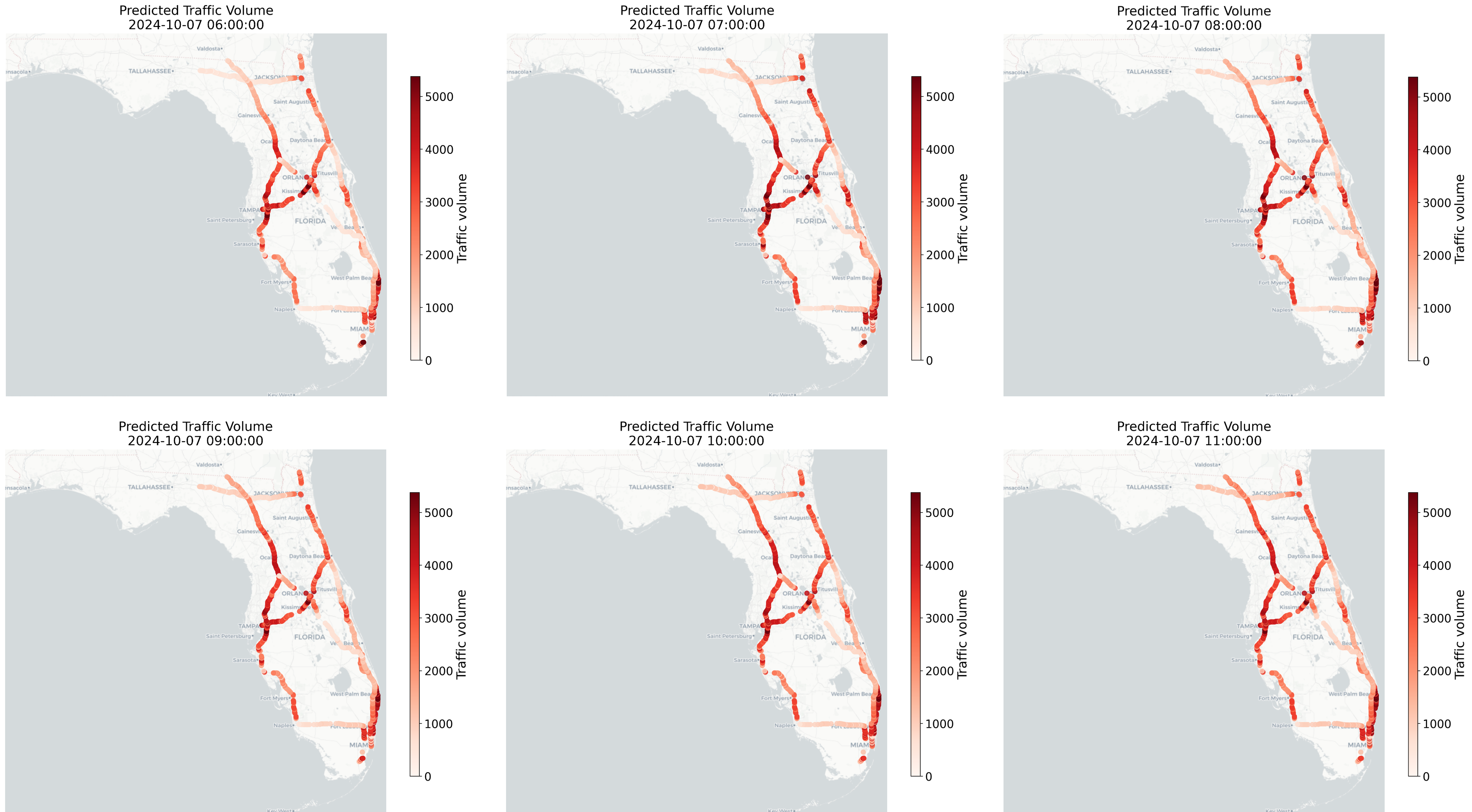}
        \caption{Congestion propagation of predicted traffic}
        \label{fig:con_predicted}
    \end{subfigure}
    \caption{Network-wide congestion propagation of traffic during Hurricane Milton}
    \label{fig:congestion}
\end{figure}

Figure~\ref{fig:congestion} compares the spatial distribution of actual and predicted traffic volumes over the 6-hour prediction horizon (6 AM to 11 AM) on October 7, 2024. Overall, the predicted traffic volume maps closely reproduce the major high-volume corridors observed in the actual traffic conditions. From 6 AM to 11 AM on October 7, 2024, the most persistent traffic concentration appears along the southeast Florida corridor, particularly near the I-95 and Florida's Turnpike facilities around the Miami--Fort Lauderdale--West Palm Beach region. The model also reproduces elevated traffic volumes  at I-4 near central Florida, as well as portions of I-75 on the west and central side of the state. Specifically, if we notice the upper north-west portion of I-75 in all distribution plots of Figure~\ref{fig:congestion}, it shows very similar congestion propagation with time for both actual and predicted traffic conditions. Overall, the locations of the main high-volume corridors and their relative intensities are well aligned with the actual observations. It demonstrates that the proposed model effectively learns spatial dependencies among detectors and preserves the relative intensity of major congestion hotspots during forecasting.

% modular training
\begin{figure}[htbp]
    \centering
    \begin{subfigure}{0.49\linewidth}
        \centering
        \includegraphics[width=\linewidth]{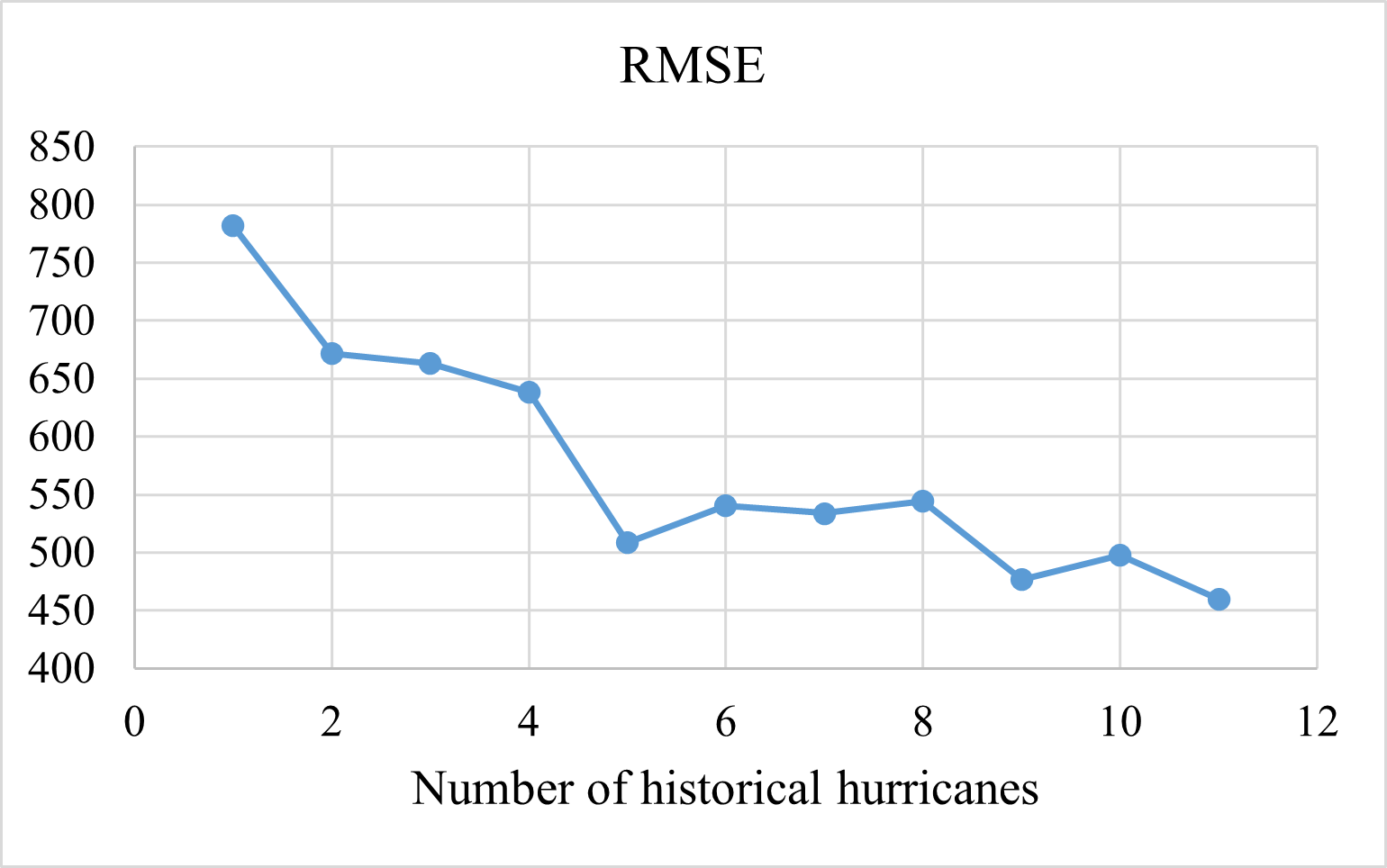}
        \caption{Changes of RMSE during Modular Training}
        \label{fig:rmse}
    \end{subfigure}
    \hfill
    \begin{subfigure}{0.49\linewidth}
        \centering
        \includegraphics[width=\linewidth]{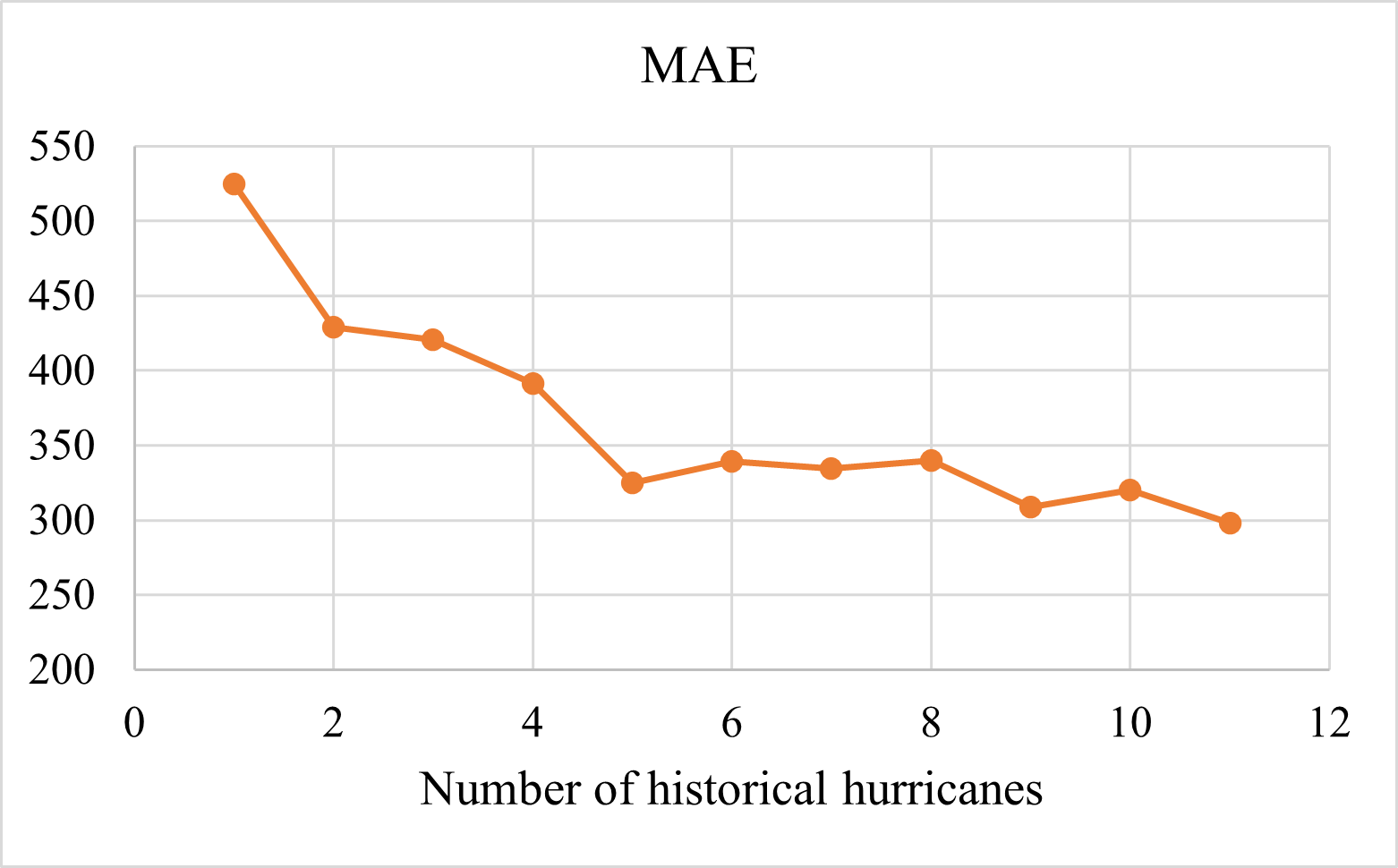}
        \caption{Changes of MAE during Modular Training}
        \label{fig:mae}
    \end{subfigure}
    \hfill
    \hfill
    \centering
    \begin{subfigure}{0.49\linewidth}
        \centering
        \includegraphics[width=\linewidth]{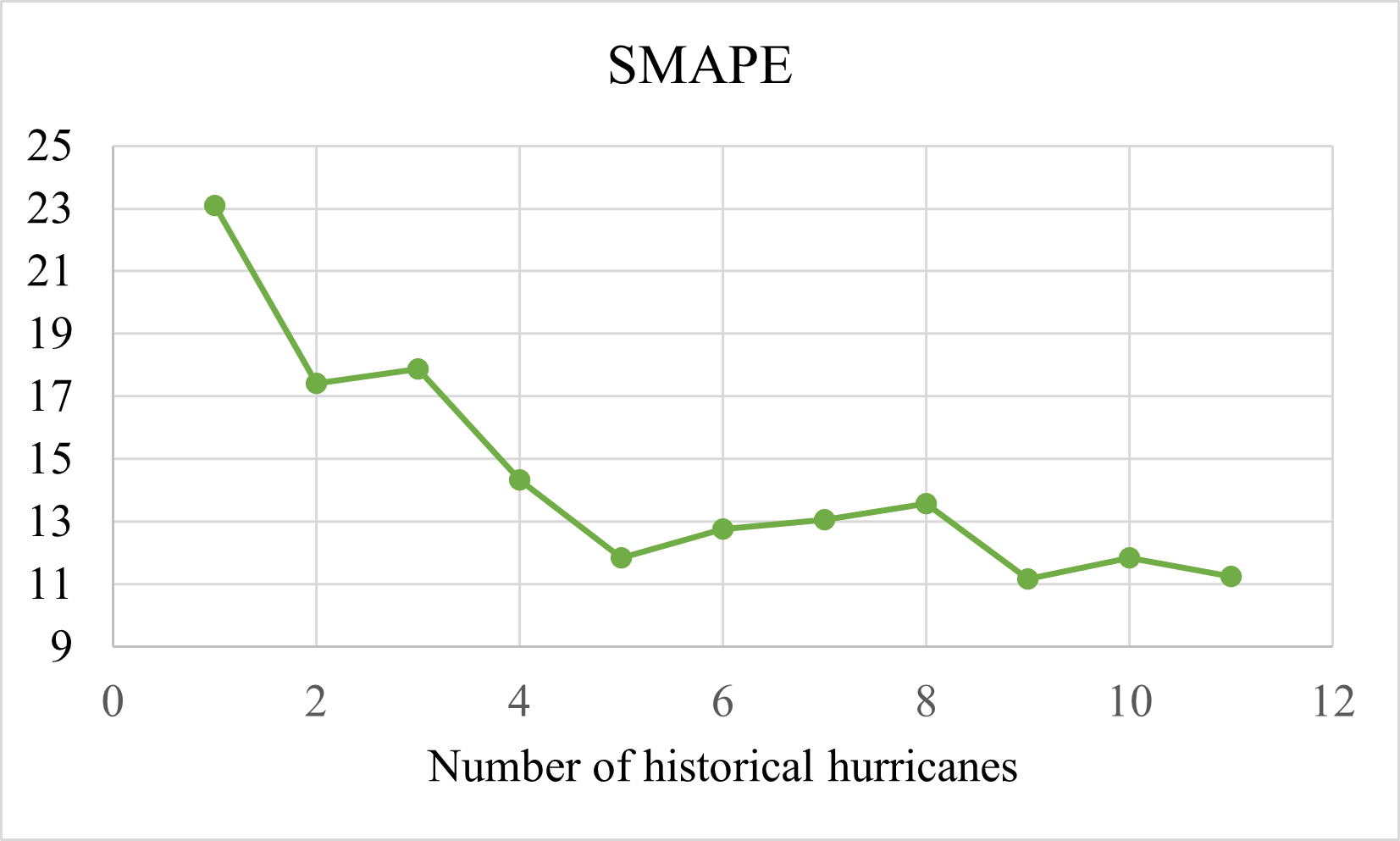}
        \caption{Changes of SMAPE during Modular Training}
        \label{fig:smape}
    \end{subfigure}
    \hfill
    \begin{subfigure}{0.49\linewidth}
        \centering
        \includegraphics[width=\linewidth]{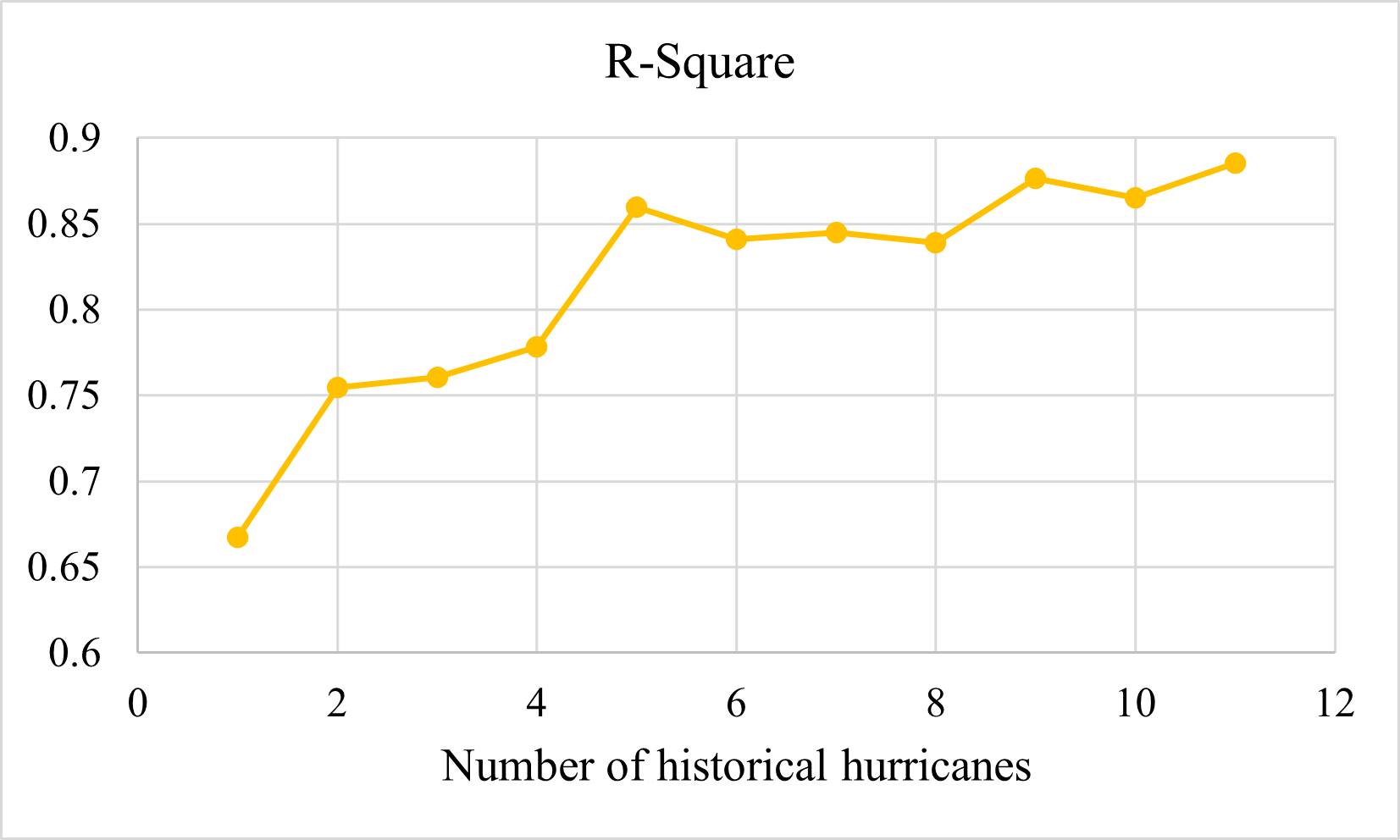}
        \caption{Changes of R\textsuperscript{2} during Modular Training}
        \label{fig:rsquare}
    \end{subfigure}
    \caption{Incremental Gain in Prediction Performance of the Generalized Dynamic Graph Learning Framework for Hurricane Milton}
    \label{fig:incremental_learning}
\end{figure}

Figure~\ref{fig:incremental_learning} shows the incremental gain of generalized dynamic graph learning model by adopting a modular training approach. We see that despite some exceptions, the model performance keeps improving as we add data from more hurricanes in the training process. Notice that there is not a significant improvement in model performance after adding data beyond from 5 or 6 hurricanes. The model performance starts to stabilize after training it with 5 to 6 immediate historical hurricanes. We can also verify this fact if we notice the prediction performance of the model for Hurricane Ian. Though the model was trained with only 6 hurricanes data before Ian, it demonstrated a very good performance for predicting the evacuation traffic of Ian. Hence, 5 to 6 immediate past hurricanes' data might be enough to make a reasonable prediction of the evacuation traffic flow during future hurricanes. 

The results do not indicate any overfitting. The model achieved good prediction performance for both Hurricane Milton and Hurricane Ian, and both hurricanes were unseen during model training. This suggests that the model can generalize evacuation traffic patterns rather than only memorizing event-specific patterns.

\section{Comparison with Baseline Models and Ablation Experiments} \label{sec:baseline_and_ablation}

To evaluate the performance of the generalized dynamic graph learning framework for evacuation traffic prediction and conduct an ablation study, we compare it against six state-of-the-art baseline models:

\begin{itemize}
    \item \textbf{LSTM}: A Long-Short Term Memory architecture designed to model temporal patterns in traffic flow. This model captures sequential dependencies but does not incorporate spatial connectivity between locations.

    \item \textbf{CNN}: A convolutional neural network architecture that captures spatial correlations using fixed-size local filters applied to the traffic feature maps. Temporal information is not explicitly modeled.

    \item \textbf{Static GCN}: A graph-based model that leverages Graph Convolutional Networks (GCNs) to model spatial dependencies over a fixed road network structure. This model does not consider temporal dynamics.

    \item \textbf{Dynamic GCN}: Similar to the Static GCN model, but the graph structure evolves over time based on the availability of detectors or changes in network connectivity. This model captures dynamic network patterns but does not consider the temporal patterns of traffic features.
    
    \item \textbf{CNN-LSTM}: A hybrid model that combines convolutional layers for extracting localized spatial features with LSTM layers to capture temporal evolution. The spatial structure is implicitly modeled through fixed-order convolutional filters.
    
    \item \textbf{Static GCN-LSTM}: A hybrid graph-based model that uses Graph Convolutional Networks (GCNs) to encode spatial dependencies based on a pre-defined, fixed road network graph. The graph structure does not change with time. The temporal dynamics are learned through LSTM layers.
\end{itemize}

% All Baseline Models and Metrics (Updated)
\begin{table}[htbp]
\centering
\caption{Baseline Model Comparison and Ablation Study for 1-hour to 6-hour prediction horizons (Milton)}
\label{tab:model_comparison}
{\scriptsize (\textit{Minimum flow = 6, Maximum flow = 10889, Mean flow = 1838, Median flow = 1503})}
{\scriptsize\begin{tabularx}{\linewidth}{l XXXXXX}
\toprule
\textbf{Model} & \textbf{Horizon} & \textbf{RMSE} & \textbf{MAE} & \textbf{SMAPE(\%)} & \textbf{R\textsuperscript{2}} & \textbf{\% Within 1.5$\sigma$} \\
\midrule
\multirow{7}{*}{LSTM *}
  & 1-hour & 390.4 & 258.1 & 11.2 & 0.92 & 93.7 \\
  & 2-hour & 472.9 & 308.1 & 11.3 & 0.88 & 93.9 \\
  & 3-hour & 528.3 & 348.4 & 11.9 & 0.85 & 93.9 \\
  & 4-hour & 561.6 & 379.0 & 13.3 & 0.83 & 93.7 \\
  & 5-hour & 595.4 & 406.6 & 14.8 & 0.81 & 93.9 \\
  & 6-hour & 650.8 & 448.0 & 16.6 & 0.76 & 94.9 \\
  & \textbf{Overall} & \textbf{539.8} & \textbf{358.0} & \textbf{13.2} & \textbf{0.84} & \textbf{93.9} \\
\midrule
\multirow{7}{*}{CNN}
  & 1-hour & 404.3 & 280.7 & 11.7 & 0.91 & 93.2 \\
  & 2-hour & 500.9 & 341.2 & 12.6 & 0.87 & 94.4 \\
  & 3-hour & 556.7 & 379.8 & 13.7 & 0.83 & 94.5 \\
  & 4-hour & 589.6 & 401.7 & 14.5 & 0.81 & 94.7 \\
  & 5-hour & 605.2 & 416.0 & 15.5 & 0.80 & 94.5 \\
  & 6-hour & 620.8 & 422.8 & 15.8 & 0.79 & 94.2 \\
  & \textbf{Overall} & \textbf{551.3} & \textbf{373.7} & \textbf{14.0} & \textbf{0.84} & \textbf{94.2} \\
\midrule
\multirow{7}{*}{Static GCN}
  & 1-hour & 505.0 & 350.4 & 14.7 & 0.87 & 93.7 \\
  & 2-hour & 555.3 & 383.3 & 15.1 & 0.84 & 93.6 \\
  & 3-hour & 601.8 & 412.5 & 16.4 & 0.81 & 93.8 \\
  & 4-hour & 628.9 & 426.1 & 16.9 & 0.79 & 94.0 \\
  & 5-hour & 626.5 & 425.8 & 17.1 & 0.78 & 93.8 \\
  & 6-hour & 627.4 & 431.6 & 18.1 & 0.78 & 93.2 \\
  & \textbf{Overall} & \textbf{592.6} & \textbf{404.9} & \textbf{16.4} & \textbf{0.81} & \textbf{93.6} \\
\midrule
\multirow{7}{*}{Dynamic GCN *}
  & 1-hour & 515.2 & 344.2 & 12.5 & 0.86 & 94.1 \\
  & 2-hour & 567.4 & 388.6 & 15.2 & 0.83 & 95.5 \\
  & 3-hour & 583.6 & 403.0 & 15.8 & 0.82 & 95.6 \\
  & 4-hour & 593.8 & 409.5 & 16.3 & 0.81 & 95.5 \\
  & 5-hour & 606.4 & 420.2 & 16.6 & 0.80 & 94.9 \\
  & 6-hour & 622.2 & 432.1 & 17.3 & 0.78 & 94.5 \\
  & \textbf{Overall} & \textbf{582.4} & \textbf{399.6} & \textbf{15.6} & \textbf{0.82} & \textbf{95.0} \\
\midrule
\multirow{7}{*}{CNN-LSTM}
  & 1-hour & 327.5 & 227.1 & 10.4 & 0.94 & 93.0 \\
  & 2-hour & 429.2 & 293.8 & 12.5 & 0.90 & 93.3 \\
  & 3-hour & 506.4 & 345.9 & 14.2 & 0.86 & 94.1 \\
  & 4-hour & 554.5 & 376.7 & 14.9 & 0.83 & 94.1 \\
  & 5-hour & 579.7 & 397.5 & 15.3 & 0.82 & 94.3 \\
  & 6-hour & 603.8 & 413.6 & 16.3 & 0.80 & 94.4 \\
  & \textbf{Overall} & \textbf{509.2} & \textbf{342.4} & \textbf{13.9} & \textbf{0.86} & \textbf{93.8} \\
\midrule
\multirow{7}{*}{Static GCN-LSTM *}
  & 1-hour & 298.2 & 198.5 & 7.7 & 0.95 & 93.1 \\
  & 2-hour & 418.0 & 273.8 & 9.6 & 0.91 & 93.2 \\
  & 3-hour & 493.1 & 326.6 & 11.9 & 0.87 & 93.5 \\
  & 4-hour & 542.2 & 360.5 & 13.5 & 0.84 & 94.0 \\
  & 5-hour & 579.2 & 385.2 & 14.7 & 0.82 & 93.9 \\
  & 6-hour & 612.4 & 402.9 & 15.7 & 0.79 & 93.9 \\
  & \textbf{Overall} & \textbf{501.9} & \textbf{324.6} & \textbf{12.2} & \textbf{0.86} & \textbf{93.5} \\
\midrule
\multirow{7}{*}{\parbox{3.1cm}{\centering \textbf{Dynamic GCN-LSTM}\\\textbf{(Proposed Model)}}}
  & 1-hour & 280.6 & 179.3 & 6.9 & 0.96 & 92.8 \\
  & 2-hour & 396.5 & 256.5 & 9.3 & 0.92 & 93.4 \\
  & 3-hour & 457.1 & 299.4 & 10.8 & 0.89 & 93.7 \\
  & 4-hour & 497.2 & 329.2 & 12.2 & 0.87 & 94.0 \\
  & 5-hour & 526.2 & 353.9 & 13.8 & 0.85 & 93.8 \\
  & 6-hour & 547.3 & 370.4 & 14.4 & 0.83 & 93.9 \\
  & \textbf{Overall} & \textbf{459.8} & \textbf{298.1} & \textbf{11.2} & \textbf{0.89} & \textbf{93.6} \\
\bottomrule
\end{tabularx}}
{\scriptsize \textit{*These models also serve as ablated version of the proposed Dynamic GCN-LSTM model.}}
\end{table}

All models were trained using the same input features, with a 6-hour input sequence to predict traffic volumes for the next 6 hours. Each models were trained for 200 epochs with a learning rate of $1 \times 10^{-5}$. The GCN-LSTM model used one GCN layer and one LSTM layer, each with 512 hidden units. Similarly, the standalone GCN and LSTM models used 512 hidden units in their respective layers. For the CNN-based models, three 1D convolutional layers were used, each with 32 filters and a kernel size of 3. 

Table~\ref{tab:model_comparison} compares the performance of various baseline models with the proposed Dynamic GCN-LSTM framework from 1-hour to 6-hour evacuation traffic prediction for Hurricane Milton. When we compare the overall results, it shows that the LSTM model, which captures only temporal dependencies, achieves an RMSE of 539.8 and an R\textsuperscript{2} score of 0.84. In contrast, models that rely solely on spatial learning such as CNN, Static GCN, and Dynamic GCN—perform relatively poorly. However, the Dynamic GCN model performs better (RMSE = 582.4, R\textsuperscript{2} = 0.82) compared to the Static GCN model (RMSE = 592.6, R\textsuperscript{2} = 0.81). This improvement highlights the value of dynamic graph representations in enhancing spatial learning by adapting to real-time changes in the traffic network. But overall, these results indicate that modeling either temporal or spatial patterns alone is insufficient for accurate evacuation traffic prediction.

Hybrid models that incorporate both spatial and temporal components demonstrate improved performance. The CNN-LSTM model yields an overall RMSE of 509.2 and R\textsuperscript{2} of 0.84, slightly better than its individual components. The Static GCN-LSTM model further improves predictive accuracy considering a fixed graph structure, bringing the overall RMSE to 501.9, while maintaining an R\textsuperscript{2} score of 0.86. The proposed Dynamic GCN-LSTM model outperforms all baseline models across every evaluation metric. It achieves the lowest overall RMSE (459.8), MAE (298.1), and SMAPE (11.2\%) values, along with the highest R\textsuperscript{2} score of 0.89 for predicting next 1-hour to 6-hour traffic flows. Compared to the Static GCN-LSTM, the Dynamic GCN-LSTM model achieves an 8.4\% improvement in overall RMSE value and an 8.2\% improvement in overall MAE value, highlighting the advantage of modeling time-varying graph structures. Also, compared to the static GCN-LSTM, the RMSE improves from 5.9\% at 1 hour to 10.6\% (65.1 veh/h) at 6 hours in the proposed model. This improvement is important because traffic management measures such as ESU and contraflow take hours to set up and must be committed to specific corridors in advance \citep{wolshon2005review1, wolshon2005review2}. With more reliable 4--6 hour advance forecasts, agencies can better decide which part of the corridors to activate and when. These results underscore the importance of capturing dynamic outage of detector-based network and validate the effectiveness of the proposed framework for evacuation traffic forecasting.

The \% of predictions within 1.5$\sigma$ remains consistently high across all baseline models, ranging between 93.0\% and 95.0\%. The proposed Dynamic GCN-LSTM model yields a slightly lower \% within (93.6\%) compared to some baselines models. This can be attributed to the proposed model's higher sensitivity and adaptability to underlying dynamics, which enables it to capture sharper demand peaks and local variations during evacuation, leading to marginally higher deviations from the mean in certain time steps, though it outperforms all baselines in terms of all other evaluation metrics. 

Compared to the baseline models, the proposed framework can provide more reliable network-wide predictions under changing detector availability and unseen hurricane conditions. This is important during hurricanes because detector outages can make the observed network structure different from the static network assumed by conventional graph learning models. Also, transportation agencies (DOTs, counties, emergency management divisions) may not have enough time or target-event data to retrain or fine-tune a model during an unfolding hurricane. Therefore, a generalized model without requiring target-event fine-tuning can be more useful for real-time operational decision making. The proposed framework can serve as a predictive decision-support tool for evacuation traffic management. Traffic flow forecasts from this framework can help agencies identify potential congestion locations, monitor bottleneck formation, prioritize traffic management resources, and decide when and where to deploy countermeasures.

Improved prediction performance becomes particularly important when traffic volumes approach roadway capacity or critical operational thresholds. During hurricane evacuations, major highways may operate near capacity \citep{StaesMenonBertini2021}, where even small changes in traffic demand can lead to substantial increases in congestion and delay \citep{dowling2004traffic}. Furthermore, traffic management measures such as emergency shoulder use (ESU) can be activated based on specific traffic volume thresholds \citep{FHWA2019DynamicShoulder}. Under such conditions, inaccurate forecasts may lead to delayed interventions or unnecessary deployment of personnel and resources. More accurate forecasts from the proposed model can help agencies better anticipate when and where traffic conditions may reach critical thresholds, potentially reducing unnecessary resource allocation and improving the timing of evacuation management strategies.

\section{Conclusions} \label{sec:conclusions}
In this study, we developed a generalized dynamic graph learning framework for predicting evacuation traffic by considering the dynamic nature of detector-based transportation networks and spatiotemporal traffic dependencies. The proposed framework integrates a dynamic graph construction approach with a GCN-LSTM model to predict evacuation traffic up to 6-hours in advance. The results demonstrate that the proposed framework can effectively utilize information from multiple historical hurricanes and provide reliable predictions for future hurricane events without requiring target-event fine-tuning. The model showed strong predictive performance for both Hurricanes Milton and Ian, suggesting that training on multiple historical hurricanes can improve model generalization. The results also show that considering evolving network topology improves prediction performance compared to static graph-based approaches. In addition, the modular training analysis showed that model performance generally improved as more historical hurricane data were added, although the improvement became smaller after including data from about 5 to 6 immediate past hurricanes. This suggests that data from a set of 5 to 6 recent historical hurricanes may be sufficient to develop a reasonable generalized prediction model for future evacuation events. The model can support transportation and emergency management agencies by providing real-time network-wide forecasts of evacuation traffic flow which can help to identify future congestion locations, and potential bottlenecks across the transportation network. These real-time forecasts can help agencies to make short-term operational decisions, such as allocating field resources, identifying critical corridors, and supporting traffic management during an unfolding evacuation event.

This study also has some limitations. The model primarily learns from historical traffic patterns and may be less adaptive to sudden external factors such as real-time weather changes, road closures, and emergency traffic management measures. For example, emergency shoulder use (ESU) was implemented during some historical hurricanes, but traffic sensors were not configured to capture vehicles using shoulder lanes. Therefore, those traffic movements were not directly represented in the dataset. In addition, the current framework is developed for evacuation traffic prediction for short-term operational decision support, not for long-term planning-level scenario analysis. Since the model relies on real-time traffic and evacuation-related information, its ability to evaluate hypothetical planning scenarios before an event is limited. Future work can develop a separate scenario-based framework to better support long-term planning-level scenario analysis. Additionally, the current framework uses distance-based edge weights that remain fixed over time. Dynamic edge representations using travel time or combined distance-travel time information may better capture evolving traffic conditions. The framework was also evaluated using Florida evacuation data only; more investigation will be required for application in other regions. Furthermore, future works may improve the framework by incorporating multimodal data sources, dynamic edge weighting strategies, real-time deployment strategies, prediction at unobserved locations, and cross-region transfer learning. The framework can also be extended for post-disaster traffic prediction to model return and re-entry traffic patterns after hurricane landfall.

\section*{Declaration of Competing Interest}
The authors declare that they have no competing financial interests or personal relationships that could have influenced the work reported in this study. A patent application has been submitted based on this research.

Given his role as Early Career Editorial Advisory Board member, Samiul Hasan had no involvement in the peer review of this article and had no access to information regarding its peer review. Full responsibility for the editorial process for this article was delegated to another journal editor.

\section*{Data Availability Statement}
The raw traffic data used in this study are owned by the Florida Department of Transportation (FDOT). With FDOT's permission, the processed data has been shared through DesignSafe Data Depot \citep{Nafees_Data_2026}. The data can be accessed from: \hyperlink{https://doi.org/10.17603/ds2-g9dd-4p21}{https://doi.org/10.17603/ds2-g9dd-4p21}.

The code supporting this study will also be archived in DesignSafe Data Depot after the paper is accepted and published.

\section*{Declaration of Generative AI}
During the preparation of this manuscript, the authors used ChatGPT for assistance in language refinement and LaTeX formatting. After using this tool/service, the authors reviewed and edited the content as needed and take full responsibility for the content of the publication.

\section*{Author Contributions Statement (CRediT)}
\textbf{M. Rafi:} Conceptualization, Data curation, Formal analysis, Investigation, Methodology, Validation, Visualization, Writing – original draft.  
\textbf{S. Hasan:} Conceptualization, Methodology, Funding acquisition, Resources, Supervision, Project administration, Validation, Writing – review and editing.  
Both authors reviewed and approved the final manuscript.

\section*{Funding}
The study was partially supported by U.S. National Science Foundation (NSF) EAGER Grant No. 2122135. However, the authors are solely responsible for the findings presented here.

\pagebreak 
\pagebreak 
\bibliography{all.bib}

\end{document}